\documentclass{article}

\usepackage[preprint]{neurips_2026}
\usepackage[utf8]{inputenc} 
\usepackage[T1]{fontenc}    
\usepackage{hyperref}       
\usepackage{url}            
\usepackage{booktabs}       
\usepackage{amsfonts}       
\usepackage{nicefrac}       
\usepackage{microtype}      
\usepackage{xcolor}         

\usepackage{graphicx}
\usepackage{amsmath}
\usepackage{booktabs} 
\usepackage{makecell}
\usepackage{mathtools}
\usepackage{ragged2e}
\usepackage{tabularx}
\usepackage{multirow} 
\usepackage{listings}

\title{VeriGeo: Controllable Geometry Question Generation with Numerical and Analytical Verification}

\author{%
Xiaoxian Duan$^{1,2}$ \quad
Zequn Liu$^{2,}$\thanks{Correspondence to: Zequn Liu  and Yingce Xia.} \quad
Yingce Xia$^{2,*}$\\
$^1$Institute of Automation, Chinese Academy of Sciences, Beijing, China\\
$^2$Zhongguancun Academy, Beijing, China\\
\texttt{duanxiaoxian2026@ia.ac.cn},\;\texttt{liuzequn@bza.edu.cn},\;\texttt{xiayingce@bza.edu.cn}
}

\newcommand{\ourM}{VeriGeo}

\begin{document}

\maketitle

\begin{abstract}
Geometry problem generation is useful for AI-assisted education and multimodal mathematical reasoning, but reliable synthesis remains difficult because the problem statement, diagram, constraints, and solution should be mutually consistent. Existing methods often trade off controllability and reliability: seed-based rewriting is flexible but weakly verifiable, whereas diagram-first construction improves validity but is less suited to arbitrary user-specified constraints.
We introduce \ourM{}, a controllable geometry generation framework grounded in executable reasoning traces. Given user constraints such as target concepts and difficulty, an Author agent generates a problem and diagram, and a Solver agent produces a proof-aligned solution. 
Both agents use a shared action sequence that connects natural language, diagrams, geometric constraints, and proof steps into a verifiable representation. A three-stage pipeline checks numerical consistency, analytical realizability, and global consistency, using verification-guided reflection to repair recoverable failures and reject unrecoverable ones. Across five LLM backbones, raw generations frequently fail these checks, while \ourM{} repairs a substantial fraction of the invalid attempts.
Supervised fine-tuning on 8.7k examples generated by \ourM{} achieves the best reported GeoQA performance among end-to-end multimodal LLM-based solvers, and obtains strong results on PGPS9K and MathVista-GPS, demonstrating the effectiveness of verified synthetic data for improving multimodal geometry reasoning. 
\end{abstract}

\section{Introduction}
Geometry serves as an important benchmark for reasoning capabilities, presenting unique challenges in both the education of students and the training of Large Language Models (LLMs)~\citep{kazemi2023geomverse,chen-etal-2021-geoqa,trinh2024alphageometry,zhang-etal-2024-geoeval}. 
Unlike general textual tasks, where the context is self-contained within the words, geometric problem-solving is inherently multimodal: it demands that a solver cross-reference textual conditions with visual constraints in a corresponding diagram~\citep{kazemi2023geomverse,chen-etal-2021-geoqa,seo2015geos,lu2021intergps}.
Consequently, high-quality geometry data is difficult to synthesize because the problem statement, diagram, symbolic constraints, and solution are expected to be mutually consistent~\citep{chen-etal-2021-geoqa,seo2015geos,lu2021intergps,fu2025trustgeogen}.
Large-scale supervision is either prohibitively expensive to curate manually or, when synthesized, suffers from compromises that degrade its reliability~\citep{fu2025trustgeogen,chen-etal-2021-geoqa,lu2021intergps,trinh2024alphageometry}. 

An ideal geometric question generator should exhibit controllability, verifiability, and diversity ~\citep{fu2025trustgeogen,zhang2025mavis}, but existing approaches typically emphasize only part of this goal. {\em Diagram-first approaches} first construct a geometric configuration, often using symbolic languages, formal graphs, or theorem-grounded construction rules, and then formulate questions based on the generated structure \citep{demoura2015lean,fu2025trustgeogen,deng2025trcot}. This improves validity, but anchoring generation to a pre-constructed diagram makes it difficult to flexibly satisfy arbitrary user-specified constraints, such as target concepts, difficulty levels, or diagram requirements \citep{singhal2014automated,fu2025trustgeogen}. {\em Seed-based generation methods} instead use LLMs to rewrite or modify existing questions \citep{yu2024metamath,zhou2023analogy,cai-etal-2024-geogpt4v}. 
They are flexible and convenient for producing variants, but their rewrite-based process can introduce hallucinated constraints and cross-modal inconsistencies among the problem statement, diagram, and solution, which are difficult to detect and repair \citep{zhou2024mathcheck,cai-etal-2024-geogpt4v}. Moreover, their diversity remains bounded by the seed distribution \citep{gao2025gllava,yu2024metamath}.

To address the aforementioned limitations, we propose \ourM{}, a generalizable agent-based framework for geometric question generation with enhanced verification capabilities. \ourM{} comprises an Author agent, which generates questions (consisting of textual descriptions and corresponding diagrams) based on user-defined constraints, and a Solver agent designed to solve them. 
Both agents operate through shared executable action sequences that connect natural language, diagrams, geometric constraints, and proof steps into verifiable representations.
Based on this executable representation, VeriGeo performs three complementary verification stages. First, numerical verification executes the action sequence and checks local geometric consistency during construction, such as whether a claimed collinearity or perpendicularity relation holds within tolerance. Second, analytical verification compiles geometric constraints into algebraic systems to test whether the configuration is geometrically realizable. Third, LLM-assisted logical verification audits global consistency among the problem text, diagram, action sequence, and solution, including contradictory assumptions, unsupported inferences, and missing cases.

Experiments across five LLM backbones show that \ourM{} substantially improves generation validity while supporting fine-grained control over difficulty and geometry concepts. 
Raw LLM generations are rarely reliable without verification, with an average direct-pass rate of only 29.02\% across backbones. Gemini-3.1-Pro, Qwen3.5-Plus, and Claude-Opus-4.6 recover 36.00\%, 30.67\%, and 20.22\% of generations through verification-guided repair, respectively, showing that repair is a major source of verified data rather than a minor post-processing step. 
In addition to improving validity, \ourM{} also broadens the conceptual coverage of generated geometry data, covering 354 distinct geometry concepts in 100 samples and surpassing both manually curated datasets and prior diagram-first or seed-based generation pipelines.
Beyond intrinsic validity, we further evaluate the usefulness of the generated data by supervised fine-tuning Qwen2.5-VL-7B-Instruct.  Training on only 8.7k verified \ourM{} examples yields strong performance on standard multimodal geometry benchmarks. Specifically, the resulting model reaches 59.40\%, 82.74\%, and 75.96\% accuracy on PGPS9K, GeoQA, and MathVista-GPS. To the best of our knowledge, \ourM{} achieves the best reported GeoQA performance among MLLM-based geometry solvers. Among prior geometry data generation methods that train MLLMs via supervised fine-tuning, \ourM{} also obtains the best reported results on PGPS9K and MathVista-GPS.

Our contribution can be summarized as follows: (1) \emph{Controllable geometry generation.} We introduce \ourM{}, a closed-loop framework that synthesizes multimodal geometry problems and proof-aligned solutions from user-specified constraints, including difficulty, target concepts, and diagram requirements. (2) \emph{Verification-guided reliability.} \ourM{} grounds generation in executable action sequences and verifies each instance through numerical, analytical, and logical checks, enabling automatic repair of invalid text--diagram--solution inconsistencies. (3) \emph{Verified data with empirical utility.} Experiments across five LLM backbones show improved generation validity and fine-grained controllability. Fine-tuning on 8.7k verified examples further yields competitive performance on standard multimodal geometry benchmarks.

\begin{figure*}[!ht]
  \centering
  \includegraphics[width=\textwidth]{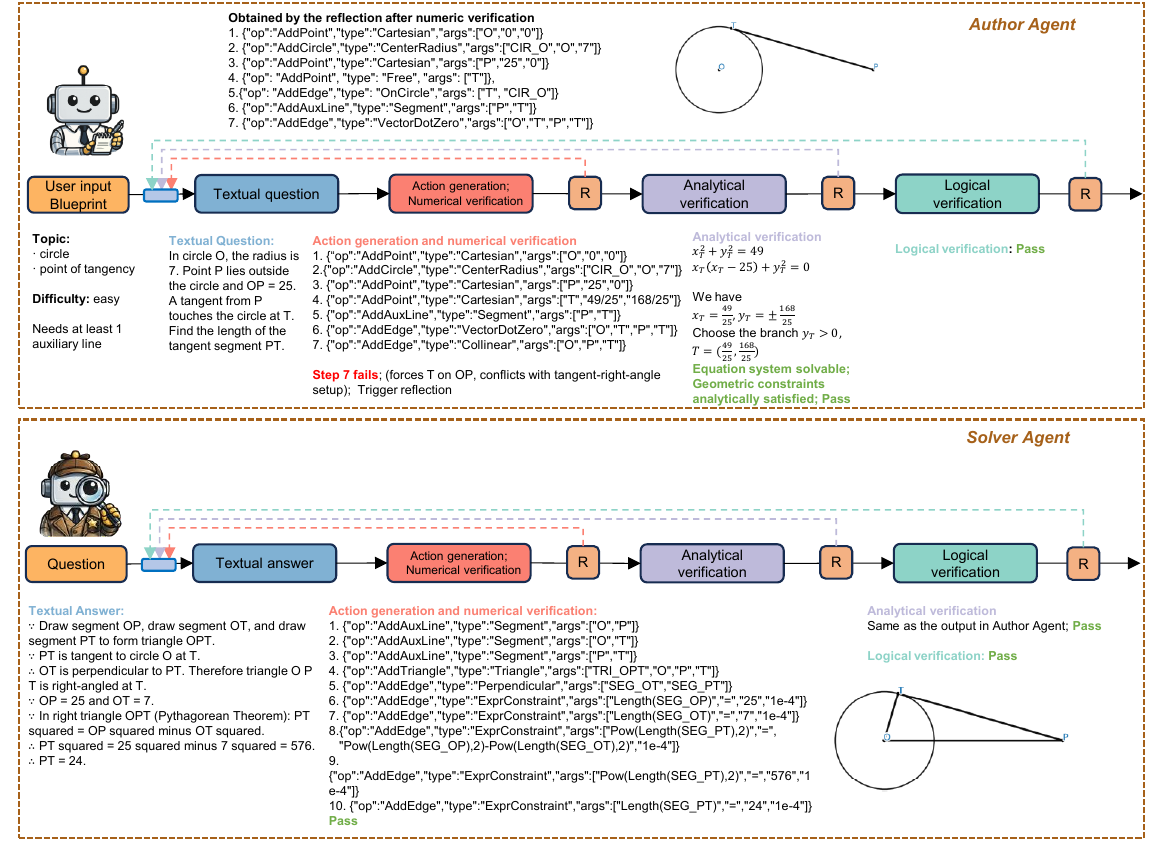}
  \caption{The workflow of \ourM{} comprises two primary components: an Author Agent for question generation and a Solver Agent for answer generation. Each agent employs a three-step verification process consisting of numerical, analytical, and logical checks. A reflection mechanism (denoted as ``R'' in the figure) is triggered specifically when verification fails. All the prompts used in this paper are provided in the supplementary materials.}
  \label{fig:verigeo_overview}
\end{figure*}

\section{Methodology}
\subsection{Overview}
We formulate geometry question generation as the synthesis of a question $Q$ and a solution $S$, subject to user-defined constraints $C$. The question $Q = (T, D)$ consists of a natural language problem statement $T$ and a corresponding diagram $D$. The constraints $C$ specify controllable attributes, such as difficulty level and required geometry concepts.

Upon receiving $C$, the framework first synthesizes a blueprint $B$ by sampling from a predefined library of geometry concepts and difficulty rules. An Author Agent, conditioned on $B$, generates the textual statement $T$ and constructs the diagram $D$ through a series of actions (denoted as $A$). To improve the correctness of the generated questions, we apply a three-step verification process: numerical, analytical, and logical verification. Any failure at these stages triggers a reflection mechanism, prompting the Author Agent to repair the content. Subsequently, a Solver Agent generates the solution $S$ based on $Q$, utilizing a process similar to that of the Author Agent.

\subsection{Blueprint Generation}
\label{sec:Blueprint Generation}
The blueprint $B$ is generated based on user constraint through LLM, which contains geometry concepts, difficulty levels, diagram complexity, etc. We maintain a manually curated library of geometry concepts spanning Euclidean geometry ($245$ geometry concepts), vector geometry ($71$ geometry concepts), and function-augmented geometry ($110$ geometry concepts). 
We consider three difficulty levels: \textsc{easy}, \textsc{medium}, and \textsc{hard}. For each difficulty level, we provide three representative examples Appendix~\ref{app:calibration-shots} and users can replace them with domain- or curriculum-specific examples when adapting \ourM{} to new educational settings. 

\subsection{Author Agent for Question Generation}
Guided by the blueprint $B$, the Author Agent first synthesizes the natural language problem statement $T$. Subsequently, it generates a sequence of executable actions to construct the corresponding diagram $D$. These action sequences undergo a three-step verification process; if any errors are detected, the agent triggers a self-reflection mechanism to refine the output.

\noindent$\rhd$ {\bf Step 1: Action generation and numerical verification.} (Full action list in Appendix \ref{app:action_catalog}.)

As illustrated in Figure \ref{fig:verigeo_overview}, the Author agent translates the text $T$ into an executable action sequence $A=\{a_t\}_t$. Each action is defined as a tuple $a_t := (\texttt{op}, \texttt{type}, \texttt{args})$, where: (1) \texttt{op} denotes the primary operation category with 11 types. For example, \texttt{AddPoint}/\texttt{AddCircle} inserts a point/circle into the diagram; \texttt{MovePoint} moves one point to specific position in the reflection stage. (2) \texttt{type} specifies the concrete method for the operation. For instance, an \texttt{AddPoint} operation may use the type \texttt{Free} (for an unconstrained point) or \texttt{Cartesian} (for a point with explicit coordinates). (3) \texttt{args} contains the parameters required to execute the specific operation. For example, a Cartesian point might require arguments such as \texttt{["P", "0", "0"]}.


Upon generating the full sequence, the system executes the script to render the diagram. This execution process inherently performs numerical verification. As each operation is processed, the system checks if the geometric constraints hold numerically against the current coordinate state. For example, as illustrated in Figure \ref{fig:verigeo_overview}, if the script asserts a constraint such as Collinear(O, P, T) but the coordinates of $T$ derived from previous steps do not lie on line $OP$ (within tolerance), the execution triggers a numerical violation error. To further improve numerical robustness, we represent generated quantities into human-readable exact forms, such as rationals (e.g., \texttt{"-2/3"}), radicals (e.g., \texttt{"\textbackslash{}sqrt(2)"}), with details in Appendix \ref{sec:digital_num}.


\noindent$\rhd$  {\bf Step 2: Analytical verification.} (Full verification list in  Appendix \ref{app:verisolve_equations}.)

Numerical verification is efficient but insufficient, as floating-point inaccuracies may allow incorrect configurations to pass within tolerance. To improve correctness, we apply analytical verification to ensure the construction is mathematically realizable.

\textit{(1) Analytic Representation.}
We first define a set of scalar variables $\mathcal{V}$ representing the diagram's degrees of freedom. Each non-fixed point $P$ is assigned variable coordinates $(x_P, y_P)$, while geometric primitives are assigned their necessary intrinsic parameters (e.g., a circle is defined by its center coordinates and radius $r$).

\textit{(2) Constraint Mapping.}
The system is constructed by iterating over the graph constraints induced by the actions in Step 1. The operation \texttt{op} could impose specific geometric constraints, which are compiled into a set of algebraic equations $\mathcal{E}$. For instance, an operation \textsf{Collinear}(A, B, C) implies that the determinant of their coordinate matrix must vanish, yielding the equation:
\begin{equation*}
(x_B-x_A)(y_C-y_A)-(y_B-y_A)(x_C-x_A)=0.
\end{equation*}

\textit{(3) Global Solving and Conflict Resolution.}
We use \texttt{sympy} \citep{meurer2017sympy} Python package to derive the analytical solution for $\mathcal{E}$. It is important to note that the resulting system $\mathcal{E}$ is frequently ill-posed: either over-determined due to redundant construction steps or under-determined due to rigid body motion invariance. To address this, we enhance the stability of the equations by implementing a series of engineering techniques, such as gauge fixing and rank-aware filtering (see Appendix~\ref{app:verisolve_equations}–-\ref{app:verisolve_closedform}).  A successful convergence produces a valid coordinate realization $\mathcal{V}^*$, certifying that the diagram is geometrically realizable. If the solver fails to converge within the specified tolerance, the instance is flagged as invalid and sent for repair.

\noindent$\rhd$ {\bf Step 3: Logical verification.}

Logical verification employs an LLM-as-a-judge paradigm to determine whether the reasoning contains logical flaws, mainly from two perspectives: (i) auditing whether the action sequence and the problem specification are logically self-consistent; and (ii) inspecting the question and solution logic for latent defects, including contradictory assumptions, invalid inferences, etc.

\noindent$\rhd$ {\bf Self-reflection}

If a failure occurs during verification, the system triggers an immediate self-reflection mechanism. By incorporating the error signals (e.g., the failing step and the reason), the module prompts the LLM to autonomously diagnose the issue and decide whether to revise the geometric question or the corresponding action sequence.

\subsection{Solver Agent for Solution Generation}
The Solver Agent synthesizes the solution $S$ for the generated question $Q=(T,D)$. This agent follows a procedure similar to that of the Author Agent: first, it generates a textual solution $S$. Next, based on $S$, it generates action sequences that must pass numerical, analytical and global verification. We consider $S$ a valid answer only if all checks pass; otherwise, a reflection step is triggered. Three specific features are worth noting: 

\noindent (1) Auxiliary Line Construction: We allow the Solver Agent to introduce auxiliary lines through dedicated operations, including but not limited to \texttt{AddAuxLine}. 

\noindent (2) Stepwise Derivation Check: For each logical step in the proof (e.g., claiming a specific angle equality or length relationship), the agent asserts a corresponding verifiable predicate. It immediately validates these claims by evaluating them against the computed coordinates. 

\noindent (3) During problem solving, the Solver may introduce additional constraints into the graph to make implicit relations explicit and to support subsequent derivations. After the full solution is constructed, we obtain an augmented graph that encodes both the original problem constraints and those added by the Solver. We then perform an analytical verification pass by traversing this augmented graph and checking global solvability/consistency.

\section{Experiments}
\subsection{Experimental Settings}
For the Author and Solver agents, we evaluated five LLM backbones: \texttt{gemini-3.1-pro-preview}, \texttt{claude-opus-4-6}, \texttt{qwen3.5-plus}, \texttt{gpt-5.4}, and \texttt{gpt-5.4-mini}. For all backbones, we set \texttt{temperature=1} and \texttt{max\_output\_tokens=60000}. We generated problems across three distinct geometric categories, including \texttt{Euclid geometry}, \texttt{Vector coordinates} and \texttt{Function calculus}. Each category is further stratified into three difficulty levels: easy, medium, and hard.  We allowed a maximum of $1$ reflection rounds whenever a verification failure is triggered for both author agent and solver agent. We conducted experiments to evaluate the generated geometric problems across three fundamental dimensions: verifiability, controllability, and diversity.

\subsection{Evaluation of Verifiability}
We first assess the efficacy of our verification mechanism in validating and repairing generated data.
To this end, we conduct a fixed-budget generation experiment with $450$ independent attempts for each LLM backbone, covering three geometric categories and three difficulty levels with $50$ attempts in each category--difficulty cell.
Unlike a quota-based setting, we do not continue generation until a predefined number of valid questions is obtained; instead, each attempt is counted once and assigned to its final verification outcome.

For each attempt, we categorize the final outcome into three mutually exclusive groups:
(1) Direct Pass, where the initial generation passes all verification checks;
(2) Repaired, where the attempt fails at least one verification stage but is successfully corrected through verification-guided reflection; and
(3) Rejected, where the attempt still fails after the maximum allowable reflection rounds.
Let $N_{\mathrm{total}}$ denote the number of generation attempts, which is $450$ for each backbone and $50$ for each category--difficulty cell.
For each outcome category $c \in \{\mathrm{Direct\ Pass}, \mathrm{Repaired}, \mathrm{Rejected}\}$, we compute $R_c = \frac{N_c}{N_{\mathrm{total}}}$, where $N_c$ is the number of attempts assigned to category $c$.


Table~\ref{tab:outcome_cost_dist} reports the distribution of verification outcomes across five LLM backbones. 
The results show that raw LLM generations remain insufficiently reliable without verification: the average \textit{Direct Pass} rate is $29.02\%$, with the weakest model, \texttt{gpt-5.4-mini}, achieving only $2.44\%$ direct pass. 
In contrast, verification-guided reflection recovers a substantial fraction of otherwise invalid attempts, with an average \textit{Repaired} rate of $25.78\%$ across backbones. 
The strongest model, \texttt{gemini-3.1-pro-preview}, achieves $54.22\%$ direct pass and further repairs $36.00\%$ of attempts, leaving only $9.78\%$ rejected. 
Similarly, \texttt{qwen3.5-plus} and \texttt{claude-opus-4-6} repair $30.67\%$ and $20.22\%$ of attempts, respectively, showing that repair is a major source of verified data rather than a minor post-processing step. 
At the same time, the high rejection rates of weaker backbones, especially \texttt{gpt-5.4-mini} with $88.00\%$ rejected and \texttt{gpt-5.4} with $58.00\%$ rejected, indicate that verification remains essential for filtering unrecoverable failures. 

Table~\ref{tab:outcome_cost_dist} also shows that verification yield varies substantially across backbones under different cost profiles. \texttt{gemini-3.1-pro-preview} achieves the highest verified yield, with only 9.78\% of attempts rejected, but its amortized cost per accepted question remains higher than that of \texttt{qwen3.5-plus} and \texttt{gpt-5.4}. By contrast, \texttt{qwen3.5-plus} provides the lowest estimated cost per verified instance in our setting, while \texttt{gpt-5.4-mini}, despite its low total cost, remains inefficient after amortization because of its high rejection rate. These results highlight the importance of accounting for verification outcomes when comparing the practical efficiency of different generation backbones.

\begin{table}[htbp]
  \centering
  \small
  \setlength{\tabcolsep}{4.8pt}
  \resizebox{0.95\textwidth}{!}{
  \begin{tabular}{lrrrrrrr}
    \toprule
    Model
    & Direct Pass & Repaired & Rejected
    & Avg. Tokens & Avg. Time (s) & Avg. Cost & Total Cost \\
    \midrule
    Gemini 3.1 Pro
      & 54.22\% & 36.00\% & 9.78\%
      & 62,733 & 449.8 & \$0.2876 & \$116.75 \\
    Qwen3.5-Plus
      & 38.22\% & 30.67\% & 31.11\%
      & 165,812 & 860.0 & \$0.0570 & \$17.66 \\
    Claude Opus 4.6
      & 40.67\% & 20.22\% & 39.11\%
      & 23,087 & 121.3 & \$0.5470 & \$149.89 \\
    GPT-5.4
      & 9.56\% & 32.44\% & 58.00\%
      & 33,624 & 79.2 & \$0.1617 & \$30.56 \\
    GPT-5.4 mini
      & 2.44\% & 9.56\% & 88.00\%
      & 34,307 & 30.6 & \$0.1691 & \$9.13 \\
    \bottomrule
  \end{tabular}
  }
  \caption{
Verification outcomes and generation costs across five LLM backbones.
Direct Pass, Repaired, and Rejected denote attempts that pass directly, pass after repair, or fail verification.
Average statistics are reported per accepted question, and Total Cost reports the total API cost.
See Table~\ref{tab:outcome_statistics_category} in Appendix~\ref{sec:additional_fig_tab} for results by geometry category and difficulty level.}
  \label{tab:outcome_cost_dist}
\end{table}

To further analyze the contribution of each verification component, we report the failure detection rate for the numerical, analytical, and logical verification modules.
Specifically, for a given module $m$, we define $N_m$ as the number of verification failure events primarily detected by module $m$, including failures that are later repaired.
Let $N_{\mathrm{acc}}$ denote the number of final accepted attempts. Failure detection rate is defined as $R_m = N_m / (N_m + N_{\mathrm{acc}}).$ Results are reported in Table \ref{tab:verification_counts}.

We can observe that the modules capture complementary failure modes. 
Numerical verification detects a large fraction of invalid attempts for most backbones, especially \texttt{gpt-5.4-mini} ($83.28\%$), \texttt{gpt-5.4} ($59.70\%$), \texttt{claude-opus-4-6} ($55.30\%$), and \texttt{qwen3.5-plus} ($54.88\%$), suggesting that many generation errors first appear as locally inconsistent geometric constructions or numerical constraints. 
Analytical verification further identifies failures that are not caught by local execution, with particularly high detection rates for \texttt{gpt-5.4-mini} ($76.32\%$) and \texttt{gpt-5.4} ($40.38\%$), highlighting the importance of checking global geometric realizability. 
Logical verification also contributes non-trivially, especially for \texttt{gemini-3.1-pro-preview} ($29.51\%$), indicating that some errors arise from higher-level inconsistencies among the problem statement, diagram, action sequence, and solution rather than from numerical or algebraic infeasibility alone. 



\begin{table*}[!t]
  \centering
  \small

  \begin{minipage}[t]{0.62\textwidth}
    \centering
    \setlength{\tabcolsep}{5pt}
    \begin{tabular}{lccc}
      \toprule
      Model & Numerical & Analytical  & Logical \\
      \midrule
      \texttt{claude-opus-4-6}
        & 55.30\% & 23.25\% & 8.36\%  \\
      \texttt{gemini-3.1-pro-preview}
        & 13.43\% & 20.08\% & 29.51\% \\
      \texttt{gpt-5.4}
        & 59.70\% & 40.38\% & 18.18\% \\
      \texttt{gpt-5.4-mini}
        & 83.28\% & 76.32\% & 10.00\% \\
      \texttt{qwen3.5-plus}
        & 54.88\% & 4.02\%  & 12.92\% \\
      \bottomrule
    \end{tabular}
    \caption{Failure detection rates of numerical, analytical, and logical verification across five LLM backbones.
Rates indicate where verification failures are first detected.
 See Table~\ref{tab:intercept-fine} in Appendix~\ref{sec:additional_fig_tab} for extended results.}
    \label{tab:verification_counts}
  \end{minipage}
  \hfill
  \begin{minipage}[t]{0.34\textwidth}
    \centering
    \setlength{\tabcolsep}{5pt}
    \begin{tabular}{lccc}
      \toprule
       & Easy & Medium & Hard \\
      \midrule
      Set 1 & 96.08\% & 93.41\% & 86.75\% \\
      Set 2 & 98.04\% & 96.70\% & 84.34\% \\
      Set 3 & 93.14\% & 95.60\% & 78.31\% \\
      \bottomrule
    \end{tabular}
    \caption{Target-difficulty matching rates (evaluated by Qwen3.5-Plus)  on generated geometry problems under three different few-shot demonstration sets.}
    \label{tab:qwen35plus_difficulty_match}
  \end{minipage}
\end{table*}

\subsection{Evaluation of Controllability}
\noindent\textbf{Controllability of difficulty level.}  
To evaluate whether the generated problems follow the intended difficulty
control, we use Qwen3.5-Plus as an independent few-shot difficulty judge.
For each generated problem, the model is asked to classify its difficulty into one of the predefined levels: easy, medium, or hard. We construct three different few-shot demonstration sets for the judge and report the
target-difficulty matching rate, i.e., the proportion of generated problems whose judged difficulty matches the intended target difficulty. Set 1 uses the same examples as those used for question generation, while Sets 2 and 3 use two independently constructed demonstration sets to test whether the evaluation is robust to the choice of few-shot examples.

As shown in Table~\ref{tab:qwen35plus_difficulty_match}, the matching rates remain consistently high across different few-shot demonstration sets. This indicates that the difficulty labels of the generated problems are largely recoverable by an independent judge and are not overly sensitive to a single choice of few-shot examples. The matching rates are especially strong for easy and medium problems, while hard problems show relatively lower agreement, suggesting that harder instances have more ambiguous or complex difficulty boundaries. We further visualize the distribution of geometric concepts across the three difficulty levels in Figure~\ref{fig:concept_count_drift} and observe that harder problems tend to involve more geometric concepts, which is consistent with the intended difficulty control and aligns with our intuition.

\paragraph{Controllability of geometry concepts.}
We further demonstrate the fine-grained control  of \ourM{} over specific geometry concepts through a case study. We curated a subset of 20 Euclidean geometry concepts and tasked the model to generate 100 problems with a strict constraint: every problem must incorporate the \textit{Pythagorean theorem}, while additional geometry concepts could be freely selected from the pool.
We then employed an LLM to extract the underlying geometry concepts from the generated outputs.

\begin{figure*}[!htbp]
  \centering
  \includegraphics[width=0.9\textwidth]{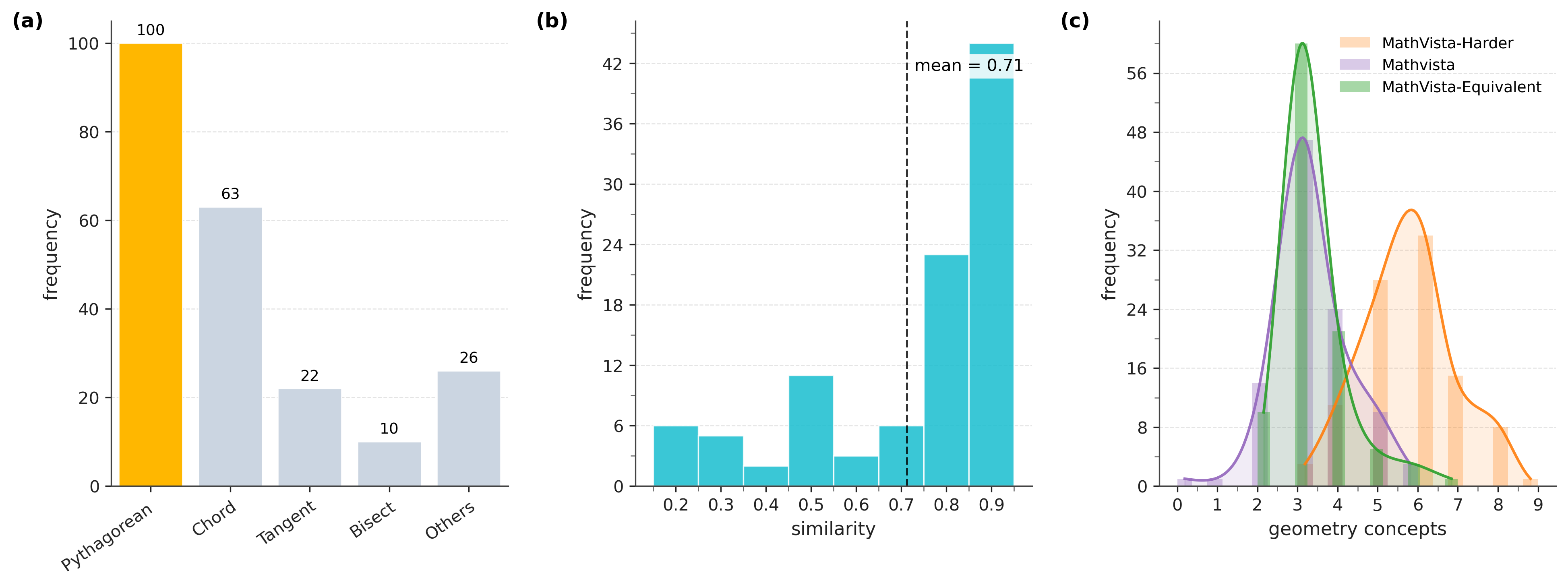}
  \caption{
  Controllability of geometry concepts and seed-conditioned generation.
  (a) geometry concept distribution under explicit concept constraints.
  (b) Knowledge similarity between generated problems and their corresponding seeds.
  (c) Distributional shift of problem complexity under difficulty-controlled,
  seed-conditioned generation on MathVista (Harder / Original / Equivalent).
  }
  \label{fig:knowledge_distributions}
\end{figure*}

As shown in Figure~\ref{fig:knowledge_distributions}(a), the Pythagorean theorem is detected in all generated questions, confirming that the model reliably follows the specified concept constraint. 
The generated problems also incorporate additional concepts from the candidate pool. In particular, center--chord perpendicularity and tangent--radius perpendicularity appear with relatively high frequency. This pattern is consistent with common pedagogical practice, where the Pythagorean theorem is often combined with circle-related perpendicularity relations to form multi-step geometry problems.

\paragraph{Generation conditioned on seed questions.}
The ability to generate new problem variants based on existing seed questions is crucial for educational applications.
To evaluate this capability, we selected 100 source problems from the MathVista \cite{lu2024MathVista} dataset and employed \ourM{} to synthesize variants conditioned on these seeds.

We first evaluate the semantic similarity between the newly generated variants and seed questions. For each seed-variant pair, we use an LLM-as-a-judge protocol to assign a knowledge-similarity score in $[0,1]$ (details in Appendix~\ref{app:seed_difficulty_modulation}). 
As shown in Figure~\ref{fig:knowledge_distributions}(b), most variants receive high scores, suggesting that \ourM{} generates new  instances while retaining the seed's  geometric knowledge.


Furthermore, we evaluate whether \ourM{} can modulate difficulty relative to a given seed question. 
For each MathVista seed, we instruct the system to generate either a harder variant or an equivalent-difficulty variant while preserving the seed's underlying geometric theme.  We use an independent pairwise difficulty-relation judge and report the target-difficulty matching rate, defined as the proportion of generated variants whose judged relation to the seed matches the requested relation (see Appendix~\ref{app:seed_difficulty_modulation} for details).  \ourM{} achieves a $100\%$ matching rate (Wilson 95\% CI \cite{WilsonCI1927}: [96.3\%, 100.0\%]) when asked to generate harder variants and an $80.0\%$ matching rate (Wilson 95\% CI: [71.1\%, 86.7\%]) when asked to generate equivalent-difficulty variants. 
This suggests that increasing difficulty is easier to control since ``preserving equivalent difficulty'' is a bit more ambiguous. 


Complementing this judge-based evaluation, we further extract the geometry concept distribution of the generated outputs in Figure~\ref{fig:knowledge_distributions}(c). 
The difficulty-increased group shifts toward higher concept counts than the original seeds, whereas the difficulty-maintained group closely follows the original distribution. 
Together, these results show that \ourM{} can control seed-conditioned problem complexity both at the judged difficulty level and at the structural level of geometry concepts, while retaining the seed's thematic context.


\subsection{Evaluation of Diversity}
To assess the diversity of our generated data, we analyze the geometry concept coverage against existing geometric question datasets. 
We randomly sampled 100 instances from \ourM{} and several baseline datasets, including: (1) manually curated datasets, \texttt{geometry3k} \cite{lu2021intergps} and \texttt{MathVista} \cite{lu2024MathVista}, (2) diagram-first approaches, \texttt{GeomVerse} \cite{kazemi2023geomverse} and \texttt{TR-GeoMM} \cite{deng2025trcot}, (3) seed-based approaches, \texttt{GeoGPT4V} \cite{cai-etal-2024-geogpt4v} and \texttt{Geo170K} \cite{gao2025gllava}. We utilized the same LLM-based extraction method used in the controllability experiments to identify the unique geometric concepts  in each sample. 

Table \ref{tab:concept_coverage} reports the total count of distinct geometry concepts covered by each dataset. 
The results demonstrate that \ourM{} encompasses a significantly broader spectrum of geometric concepts compared to prior works. The diversity of diagram-first approaches is constrained by the the space of reliably renderable diagrams. Seed-based approaches expand the coverage by leveraging LLMs, yet their diversity is still bounded by the concept distribution of the seed questions.
Notably, \ourM{} achieves the highest concept coverage, surpassing even the manually curated datasets, suggesting that our framework can systematically explore a wider concept space, thereby offering better alignment with real-world mathematical education requirements.

\begin{table*}[t]
  \centering
  \small
  \setlength{\tabcolsep}{5pt}
  \resizebox{0.95\textwidth}{!}{
  \begin{tabular}{c cc cc cc c}
    \toprule
    & \multicolumn{2}{c}{Manually Curated}
    & \multicolumn{2}{c}{Diagram-First Approaches}
    & \multicolumn{2}{c}{Seed-based Generation}
    & Ours \\
    \cmidrule(lr){2-3}\cmidrule(lr){4-5}\cmidrule(lr){6-7}\cmidrule(lr){8-8}
    & \texttt{geometry3k} & \texttt{MathVista}
    & \texttt{GeomVerse} & \texttt{TR-GeoMM}
    & \texttt{GeoGPT4V} & \texttt{Geo170K}
    & \texttt{VeriGeo} \\
    \midrule
    \# Geometry concept & 169 & 201 & 116 & 126 & 177 & 187 & 354 \\
    \bottomrule
  \end{tabular}
  }
  \caption{\centering Comparison of geometry concept coverage across datasets.}
  \label{tab:concept_coverage}
\end{table*}

\subsection{Evaluation of downstream training}
To evaluate whether the data generated by \ourM{} is useful beyond question generation, we further use it for downstream supervised fine-tuning.  We fine-tune \texttt{Qwen2.5-VL-7B-Instruct} on all verified examples generated by \ourM{} (8.7k instances in total).
We focus on supervised fine-tuning and leave reinforcement learning or process-level optimization for future work. 

We evaluate the fine-tuned model on three standard multimodal geometry reasoning benchmarks:
PGPS9K~\cite{Zhang2023PGPS}, GeoQA~\cite{chen-etal-2021-geoqa}, and MathVista-GPS~\cite{lu2024MathVista}.
The results are reported in Table~\ref{tab:downstream_training}. Fine-tuning on the verified \ourM{} data substantially improves the base model across all three benchmarks. 
Compared with \texttt{Qwen2.5-VL-7B-Instruct}, \ourM{} achieving $59.40\%$, $82.74\%$, and $75.96\%$ on the three benchmarks, respectively.  
On GeoQA, \ourM{} achieves the best reported result among end-to-end MLLM-based geometry solvers.We note that prior neural-symbolic systems~\citep{zhang2025fgeohypergnet,ping2026autogps}
report higher GeoQA scores, but they rely on formal symbolic systems, external formalization,
or deductive reasoning engines, and are therefore not directly comparable to our setting. On PGPS9K and MathVista-GPS, \ourM{} also obtains the best reported results among prior geometry data generation methods that train MLLMs via supervised fine-tuning.
Notably, these gains are obtained with only 8.7k verified training examples, substantially fewer than prior geometry data-generation methods such as G-LLaVA, MAVIS, TR-CoT, and GeoGen-SFT, which typically use tens or hundreds of thousands of examples.  These results suggest that the verified data generated by \ourM{} provides effective supervision for multimodal geometry reasoning, and that data quality and verifiability can be as important as data scale.

\begin{table}[!htbp]
\centering
\small
\setlength{\tabcolsep}{4.2pt}
\caption{
Supervised fine-tuning of Qwen2.5-VL-7B-Instruct on 8.7k verified examples generated by \ourM{}.
All numbers are accuracy (\%). 
``--'' indicates unavailable results under the same benchmark protocol. Data Scale refers to task-specific math/geometry SFT data when available. See Table \ref{tab:downstream_training_ext} in Appendix \ref{sec:additional_fig_tab} for an extended comparison with additional baselines and model tuning parameters.
}
\label{tab:downstream_training}
\resizebox{\textwidth}{!}{
\begin{tabular}{lccccc}
\toprule
\textbf{Model} & \textbf{Params} & \textbf{PGPS9K} & \textbf{GeoQA} & \textbf{MathVista-GPS} & \textbf{Data Scale} \\
\midrule
\multicolumn{6}{l}{\textit{Open-source general MLLMs}} \\
InternVL2-76B~\citep{wang2025mathcodervl} & 76B & -- & -- & 67.80 & -- \\
Qwen2.5-VL-7B-Instruct & 7B & 42.40 & 72.07 & 53.37 & -- \\
\midrule
\multicolumn{6}{l}{\textit{Open-source math-tuned MLLMs}} \\
Math-LLaVA-13B~\citep{shi2024mathllava,wang2025mathcodervl} & 13B & -- & 47.80 & 57.70 & 360K \\
MathGLM-Vision-9B~\citep{yang2024mathglm} & 9B & -- & -- & 64.42 & MathVL + VQA \\
Math-PUMA-Qwen2~\citep{wang2025mathcodervl} & 7B/8B & -- & -- & 48.10 & -- \\
MathCoder-VL-8B~\citep{wang2025mathcodervl} & 8B & -- & -- & 73.60 & 8.6M + 3M \\
\midrule
\multicolumn{6}{l}{\textit{SFT-only geometry data-generation methods}} \\
G-LLaVA-7B~\citep{gao2025gllava,deng2025trcot} & 7B & -- & 62.80 & 53.40 & 170K \\
G-LLaVA-13B~\citep{gao2025gllava,pan2025geogen} & 13B & -- & 67.00 & 56.70 & 170K \\
MAVIS-7B~\citep{zhang2025mavis,pan2025geogen} & 7B & -- & 68.30 & 64.10 & 834K \\
TR-CoT-8B~\citep{deng2025trcot,pan2025geogen} & 8B & -- & 75.90 & 73.10 & 87K \\
GeoGen-SFT-3B~\citep{pan2025geogen} & 3B & 44.70 & 75.60 & 64.50 & 224K \\
GeoGen-SFT-7B~\citep{pan2025geogen} & 7B & 54.30 & 78.00 & 74.00 & 224K \\
TR-CoT-Qwen2.5-VL-7B~\citep{deng2025trcot} & 7B & -- & 79.20 & --$^\dagger$ & 65K + Geo170K \\
TR-CoT-InternVL2.5-8B~\citep{deng2025trcot} & 8B & -- & 76.70 & --$^\dagger$ & 65K + Geo170K \\
\midrule
\textbf{VeriGeo} & 7B & \textbf{59.40} & \textbf{82.74} & \textbf{75.96} & \textbf{8.7k} \\
\bottomrule
\end{tabular}
}
\begin{flushleft}
\footnotesize
$^\dagger$ TR-CoT reports MathVista results under its own geometry-problem evaluation setting, but does not explicitly report the exact MathVista-GPS split used in our evaluation; therefore, we do not copy those numbers into the MathVista-GPS column. 
\end{flushleft}
\end{table}

\section{Related Work}
\label{sec:related_work}
We organize prior work into \emph{Diagram-First Approaches} and \emph{Seed-based Generation}.


\noindent{\bf Diagram-First Approaches.}
Diagram-first methods treat the geometric configuration as the source of truth and derive text and solutions from that structured state. Early educational generators enumerate figures, instantiate templates, and validate solvability via automated deduction \citep{singhal2014automated}. Recent work strengthens the substrate and verification loop: GeomVerse uses procedural construction for controlled evaluation \citep{kazemi2023geomverse}, while MAVIS synthesizes diagram--text--reasoning triples at scale \citep{zhang2025mavis}. Theorem/constraint-grounded pipelines further tighten correctness, e.g., TR-CoT and TrustGeoGen \citep{deng2025trcot,fu2025trustgeogen}, and unified generators such as GeoUni aim to jointly model diagram, problem text, and solutions under concept control \citep{cheng2025geouni}. Symbolic-first solvers are also diagram-first in spirit: AlphaGeometry operates over a formal language and couples synthetic theorem/proof generation with sound deduction \citep{trinh2024alphageometry}. Diagram-first pipelines provide strong \emph{verifiability}, but their \emph{diversity} and \emph{controllability} are often constrained by engineered primitives and template libraries.

\noindent{\bf Seed-based Generation.}
Seed-based generation expands datasets by rewriting or evolving existing problems, enabling fast linguistic diversification with limited domain engineering. General recipes include Self-Instruct and Evol-Instruct-style mutation \citep{wang2023selfinstruct,xu2023wizardlm}, with math-specialized variants such as WizardMath reinforcing multi-step reasoning \citep{luo2025wizardmath}. In multimodal learning, Visual Instruction Tuning (e.g., LLaVA) similarly scales supervision from image--text seeds \citep{liu2023visual}. In geometry, GeoGPT4V and G-LLaVA build large corpora of diagram-aligned problems \citep{cai-etal-2024-geogpt4v,gao2025gllava}, while GeoThought emphasizes step-wise reasoning traces (and reflection) \citep{shi2025geothought}. Seed-based generation is also used to synthesize \emph{harder evaluation} beyond static benchmarks, as in CHASE~\citep{patel2025challenging} and robustness transformations such as MathCheck \citep{zhou2024mathcheck}. This paradigm offers convenient \emph{control} via targeted edits, but its distribution is anchored to the seed corpus; moreover, without an executable representation, it has weaker \emph{verifiability} and is vulnerable to hallucinations and silent solvability/text--diagram inconsistencies.

\section{Conclusions}
We introduced \ourM{}, a controllable framework for generating multimodal geometry problems with verified diagrams and proof-aligned solutions.  \ourM{} grounds both question generation and solution generation in a shared executable action sequence, allowing natural-language statements, diagrams, geometric constraints, and reasoning steps to be checked within a unified representation. 
Experiments across five LLM backbones show that raw generations are often unreliable, while verification-guided repair substantially increases the yield of valid problems. The generated data also exhibits fine-grained controllability over target difficulty and geometry concepts, while covering a broader range of concepts than prior manually curated, diagram-first, or seed-based datasets. 
Beyond intrinsic generation quality, supervised fine-tuning on only 8.7k verified \ourM{} examples leads to strong performance on standard multimodal geometry benchmarks, including the best reported GeoQA result among end-to-end multimodal LLM-based solvers.

\bibliography{reference}
\bibliographystyle{abbrv}

\clearpage
\appendix
\setcounter{table}{0}
\setcounter{figure}{0}
\renewcommand{\thetable}{A\arabic{table}}
\renewcommand{\thefigure}{A\arabic{figure}}

\section{Limitations and Future Work}
\label{app:limitations_future_work}

VeriGeo provides a controllable and verifiable framework for generating multimodal geometry problems, but several directions remain open for future extension.

First, the current implementation focuses on geometry problems that can be represented by our executable action grammar and verified through numerical, analytical, and logical checks. This design gives VeriGeo strong controllability and enables automatic rejection of inconsistent samples, but the supported problem space is still tied to the available geometric operators, constraint types, and algebraic verification routines. Extending the action grammar to cover richer diagram types, more advanced theorem-level constructions, three-dimensional geometry, and curriculum-specific variants would further broaden the applicability of the framework.

Second, the generation efficiency of VeriGeo depends on the choice of backbone model and the verification--reflection budget. Our experiments show that different backbones exhibit different direct-pass, repair, rejection, time, token, and cost profiles. This suggests several practical extensions, including adaptive model routing, early rejection of unrecoverable generations, dynamic reflection budgets, and lightweight verifier-based pre-filtering before invoking stronger models.

Finally, our downstream evaluation focuses on supervised fine-tuning with verified VeriGeo data. The strong gains obtained from a relatively small verified dataset indicate that verification quality is an important factor for multimodal geometry reasoning. Future work can further explore process-level supervision, reinforcement learning, verifier-guided data selection, and larger-scale training across different multimodal backbones. Beyond geometry, the same principle of executable generation followed by multi-stage verification may also be extended to other structured multimodal reasoning domains, such as physics diagrams, charts, and symbolic visual reasoning.

\subsection*{Usage of Large Language Models}
Large language models and agentic workflows are part of the proposed method in this paper. 
Specifically, \ourM{} uses LLMs as Author and Solver agents for geometry problem generation and solution generation. 
LLMs are also used for verification-guided reflection and for LLM-as-a-judge evaluations, including difficulty classification, pairwise difficulty comparison, geometry-concept extraction, and knowledge-similarity scoring. 
These uses are core components of the framework and are described in the methodology and experimental setup sections.

The authors are responsible for all scientific content in the paper, including the method design, implementation, dataset construction, experimental design, result analysis, figures, tables, and references. 
All LLM-generated outputs used in the framework were subject to the verification and evaluation procedures described in the paper. 
During manuscript preparation, LLMs were used only for auxiliary editing support, such as grammar checking, stylistic polishing, and improving clarity and fluency. 
No LLM or agent was treated as an author.

\clearpage

\section{Additional Tables and Figures}
\label{sec:additional_fig_tab}
\begin{table}[!htbp]
\centering
\caption{\textbf{Fine-grained outcome distribution and generation cost} (extension of Table~\ref{tab:outcome_cost_dist}). For every authoring model we report the percentage of \textit{Direct Pass}, \textit{Accepted after Repair}, and \textit{Rejected} attempts on the $n=50$ problems in each (domain, difficulty) cell. Avg. Tokens, Avg. Time, and Avg. Cost are reported per accepted problem in each cell, where the number of accepted problems is computed as $50 \times (\textit{Direct Pass}+\textit{Repaired})$. Total Cost reports the total API cost incurred in the corresponding cell and sums to the model-level total. The \textit{Overall} row aggregates the nine cells of each model and reproduces the model-level numbers reported in the main text.}
\label{tab:outcome_statistics_category}
\scriptsize
\setlength{\tabcolsep}{3pt}
\resizebox{\textwidth}{!}{
\begin{tabular}{llccccrrrr}
\toprule
\textbf{Model} & \textbf{Domain} & \textbf{Difficulty}
& \textbf{Direct Pass\,(\%)} & \textbf{Repaired\,(\%)} & \textbf{Rejected\,(\%)}
& \textbf{Avg. Tokens} & \textbf{Avg. Time (s)} & \textbf{Avg. Cost} & \textbf{Total Cost} \\
\midrule
\multirow{9}{*}{Gemini-3.1-Pro}
& \multirow{3}{*}{Euclid.} & Easy   & 48.0 & 40.0 & 12.0 & 70,908 & 477.6 & \$0.3266 & \$14.37 \\
&  & Medium & 46.0 & 32.0 & 22.0 & 80,295 & 559.8 & \$0.4395 & \$17.14 \\
&  & Hard   & 32.0 & 26.0 & 42.0 & 77,074 & 462.1 & \$0.5586 & \$16.20 \\
\cmidrule(l){2-10}
& \multirow{3}{*}{Func.\,\&\,Calc.} & Easy   & 70.0 & 28.0 & 2.0 & 47,040 & 318.8 & \$0.1735 & \$8.50 \\
&  & Medium & 64.0 & 34.0 & 2.0 & 50,714 & 342.9 & \$0.1963 & \$9.62 \\
&  & Hard   & 64.0 & 34.0 & 2.0 & 48,035 & 340.0 & \$0.1814 & \$8.89 \\
\cmidrule(l){2-10}
& \multirow{3}{*}{Vec.\,\&\,Coord.} & Easy   & 56.0 & 44.0 & 0.0 & 60,657 & 399.0 & \$0.2432 & \$12.16 \\
&  & Medium & 56.0 & 42.0 & 2.0 & 69,486 & 569.2 & \$0.2994 & \$14.67 \\
&  & Hard   & 52.0 & 44.0 & 4.0 & 70,865 & 613.8 & \$0.3167 & \$15.20 \\
\cmidrule(l){2-10}
& \multicolumn{2}{r}{\textit{Overall}} & \textbf{54.22} & \textbf{36.00} & \textbf{9.78}
& \textbf{62,733} & \textbf{449.8} & \textbf{\$0.2876} & \textbf{\$116.75} \\
\midrule

\multirow{9}{*}{Qwen3.5-Plus}
& \multirow{3}{*}{Euclid.} & Easy   & 48.0 & 38.0 & 14.0 & 160,465 & 842.3 & \$0.0456 & \$1.96 \\
&  & Medium & 44.0 & 36.0 & 20.0 & 132,197 & 851.4 & \$0.0438 & \$1.75 \\
&  & Hard   & 36.0 & 36.0 & 28.0 & 174,101 & 997.1 & \$0.0594 & \$2.14 \\
\cmidrule(l){2-10}
& \multirow{3}{*}{Func.\,\&\,Calc.} & Easy   & 28.0 & 26.0 & 46.0 & 179,265 & 730.3 & \$0.0674 & \$1.82 \\
&  & Medium & 20.0 & 20.0 & 60.0 & 194,960 & 772.2 & \$0.1000 & \$2.00 \\
&  & Hard   & 22.0 & 20.0 & 58.0 & 208,910 & 847.3 & \$0.1038 & \$2.18 \\
\cmidrule(l){2-10}
& \multirow{3}{*}{Vec.\,\&\,Coord.} & Easy   & 58.0 & 30.0 & 12.0 & 156,191 & 858.3 & \$0.0425 & \$1.87 \\
&  & Medium & 46.0 & 36.0 & 18.0 & 155,969 & 887.8 & \$0.0468 & \$1.92 \\
&  & Hard   & 42.0 & 34.0 & 24.0 & 172,440 & 876.9 & \$0.0532 & \$2.02 \\
\cmidrule(l){2-10}
& \multicolumn{2}{r}{\textit{Overall}} & \textbf{38.22} & \textbf{30.67} & \textbf{31.11}
& \textbf{165,812} & \textbf{860.0} & \textbf{\$0.0570} & \textbf{\$17.66} \\
\midrule

\multirow{9}{*}{Claude-Opus-4.6}
& \multirow{3}{*}{Euclid.} & Easy   & 38.0 & 18.0 & 44.0 & 21,058 & 105.5 & \$0.5061 & \$14.17 \\
&  & Medium & 30.0 & 14.0 & 56.0 & 26,426 & 150.0 & \$0.8536 & \$18.78 \\
&  & Hard   & 26.0 & 12.0 & 62.0 & 27,420 & 155.1 & \$1.0311 & \$19.59 \\
\cmidrule(l){2-10}
& \multirow{3}{*}{Func.\,\&\,Calc.} & Easy   & 56.0 & 28.0 & 16.0 & 18,094 & 84.4  & \$0.2850 & \$11.97 \\
&  & Medium & 52.0 & 28.0 & 20.0 & 20,953 & 105.1 & \$0.3588 & \$14.35 \\
&  & Hard   & 50.0 & 24.0 & 26.0 & 23,192 & 122.5 & \$0.4514 & \$16.70 \\
\cmidrule(l){2-10}
& \multirow{3}{*}{Vec.\,\&\,Coord.} & Easy   & 44.0 & 22.0 & 34.0 & 20,672 & 95.4  & \$0.3955 & \$13.05 \\
&  & Medium & 34.0 & 18.0 & 48.0 & 26,204 & 148.2 & \$0.7431 & \$19.32 \\
&  & Hard   & 36.0 & 18.0 & 46.0 & 30,152 & 176.2 & \$0.8133 & \$21.96 \\
\cmidrule(l){2-10}
& \multicolumn{2}{r}{\textit{Overall}} & \textbf{40.67} & \textbf{20.22} & \textbf{39.11}
& \textbf{23,087} & \textbf{121.3} & \textbf{\$0.5470} & \textbf{\$149.89} \\
\midrule

\multirow{9}{*}{GPT-5.4}
& \multirow{3}{*}{Euclid.} & Easy   & 6.0 & 24.0 & 70.0 & 31,907 & 68.2  & \$0.2240 & \$3.36 \\
&  & Medium & 4.0 & 18.0 & 78.0 & 36,378 & 105.3 & \$0.3464 & \$3.81 \\
&  & Hard   & 6.0 & 18.0 & 76.0 & 35,430 & 108.2 & \$0.3192 & \$3.83 \\
\cmidrule(l){2-10}
& \multirow{3}{*}{Func.\,\&\,Calc.} & Easy   & 12.0 & 42.0 & 46.0 & 32,205 & 65.0 & \$0.1078 & \$2.91 \\
&  & Medium & 12.0 & 38.0 & 50.0 & 33,696 & 68.6 & \$0.1268 & \$3.17 \\
&  & Hard   & 12.0 & 38.0 & 50.0 & 35,812 & 64.0 & \$0.1344 & \$3.36 \\
\cmidrule(l){2-10}
& \multirow{3}{*}{Vec.\,\&\,Coord.} & Easy   & 12.0 & 38.0 & 50.0 & 30,535 & 96.0 & \$0.1300 & \$3.25 \\
&  & Medium & 10.0 & 36.0 & 54.0 & 31,034 & 87.4 & \$0.1387 & \$3.19 \\
&  & Hard   & 12.0 & 40.0 & 48.0 & 37,181 & 77.3 & \$0.1415 & \$3.68 \\
\cmidrule(l){2-10}
& \multicolumn{2}{r}{\textit{Overall}} & \textbf{9.56} & \textbf{32.44} & \textbf{58.00}
& \textbf{33,624} & \textbf{79.2} & \textbf{\$0.1617} & \textbf{\$30.56} \\
\midrule

\multirow{9}{*}{GPT-5.4-mini}
& \multirow{3}{*}{Euclid.} & Easy   & 2.0 & 10.0 & 88.0 & 32,709 & 26.8 & \$0.1717 & \$1.03 \\
&  & Medium & 4.0 & 12.0 & 84.0 & 35,527 & 32.6 & \$0.1288 & \$1.03 \\
&  & Hard   & 2.0 & 14.0 & 84.0 & 36,492 & 34.2 & \$0.1375 & \$1.10 \\
\cmidrule(l){2-10}
& \multirow{3}{*}{Func.\,\&\,Calc.} & Easy   & 2.0 & 4.0 & 94.0 & 42,378 & 34.1 & \$0.4033 & \$1.21 \\
&  & Medium & 2.0 & 8.0 & 90.0 & 38,465 & 33.1 & \$0.2260 & \$1.13 \\
&  & Hard   & 0.0 & 2.0 & 98.0 & 32,938 & 23.8 & \$0.8000 & \$0.80 \\
\cmidrule(l){2-10}
& \multirow{3}{*}{Vec.\,\&\,Coord.} & Easy   & 4.0 & 18.0 & 78.0 & 32,152 & 30.5 & \$0.0855 & \$0.94 \\
&  & Medium & 2.0 & 10.0 & 88.0 & 31,112 & 27.1 & \$0.1550 & \$0.93 \\
&  & Hard   & 4.0 & 8.0 & 88.0 & 31,242 & 28.1 & \$0.1600 & \$0.96 \\
\cmidrule(l){2-10}
& \multicolumn{2}{r}{\textit{Overall}} & \textbf{2.44} & \textbf{9.56} & \textbf{88.00}
& \textbf{34,307} & \textbf{30.6} & \textbf{\$0.1691} & \textbf{\$9.13} \\
\bottomrule
\end{tabular}
}
\end{table}

\clearpage

\clearpage
\begin{table}[!htbp]
\centering
\caption{\textbf{Fine-grained failure detection rates} (extension of Table~\ref{tab:verification_counts}).
For each (model, domain, difficulty) cell, we report the failure detection rate
$R_m = N_m/(N_m+N_{\mathrm{acc}})$ for each verification stage
$m\in\{\text{Numerical}, \text{Analytical}, \text{Logical}\}$.
Here, $N_m$ denotes the number of failure cases first detected by stage $m$,
including cases that are later repaired, and $N_{\mathrm{acc}}$ denotes the number of final accepted attempts.
The \textit{Overall} row pools the nine cells of each model and reproduces the model-level
rates reported in the main text.}
\label{tab:intercept-fine}
\scriptsize
\setlength{\tabcolsep}{5pt}
\begin{tabular}{llcccc}
\toprule
\textbf{Model} & \textbf{Domain} & \textbf{Difficulty} & $R_{\text{Num.}}$\,(\%) & $R_{\text{Ana.}}$\,(\%) & $R_{\text{Log.}}$\,(\%) \\
\midrule
\multirow{9}{*}{Gemini-3.1-Pro} & \multirow{3}{*}{Euclid.} & Easy & 0.0 & 48.2 & 0.0 \\
 &  & Medium & 0.0 & 45.8 & 0.0 \\
 &  & Hard & 0.0 & 48.2 & 0.0 \\
\cmidrule(l){2-6}
 & \multirow{3}{*}{Func.\,\&\,Calc.} & Easy & 24.6 & 0.0 & 23.4 \\
 &  & Medium & 19.7 & 0.0 & 31.9 \\
 &  & Hard & 21.0 & 0.0 & 32.9 \\
\cmidrule(l){2-6}
 & \multirow{3}{*}{Vec.\,\&\,Coord.} & Easy & 12.3 & 0.0 & 42.5 \\
 &  & Medium & 14.0 & 2.0 & 41.0 \\
 &  & Hard & 12.7 & 0.0 & 43.5 \\
\cmidrule(l){2-6}
 & \multicolumn{2}{r}{\textit{Overall}} & \textbf{13.4} & \textbf{20.1} & \textbf{29.5} \\
\midrule
\multirow{9}{*}{Qwen3.5-Plus} & \multirow{3}{*}{Euclid.} & Easy & 46.2 & 6.5 & 14.0 \\
 &  & Medium & 49.4 & 4.8 & 18.4 \\
 &  & Hard & 52.6 & 5.3 & 18.2 \\
\cmidrule(l){2-6}
 & \multirow{3}{*}{Func.\,\&\,Calc.} & Easy & 62.0 & 3.6 & 12.9 \\
 &  & Medium & 71.0 & 0.0 & 4.8 \\
 &  & Hard & 69.1 & 4.5 & 0.0 \\
\cmidrule(l){2-6}
 & \multirow{3}{*}{Vec.\,\&\,Coord.} & Easy & 46.3 & 2.2 & 12.0 \\
 &  & Medium & 51.2 & 0.0 & 12.8 \\
 &  & Hard & 51.3 & 7.3 & 11.6 \\
\cmidrule(l){2-6}
 & \multicolumn{2}{r}{\textit{Overall}} & \textbf{54.9} & \textbf{4.0} & \textbf{12.9} \\
\midrule
\multirow{9}{*}{Claude-Opus-4.6} & \multirow{3}{*}{Euclid.} & Easy & 52.5 & 34.9 & 9.7 \\
 &  & Medium & 57.7 & 47.6 & 0.0 \\
 &  & Hard & 65.5 & 42.4 & 0.0 \\
\cmidrule(l){2-6}
 & \multirow{3}{*}{Func.\,\&\,Calc.} & Easy & 48.1 & 0.0 & 20.8 \\
 &  & Medium & 54.0 & 2.4 & 4.8 \\
 &  & Hard & 53.2 & 7.5 & 11.9 \\
\cmidrule(l){2-6}
 & \multirow{3}{*}{Vec.\,\&\,Coord.} & Easy & 52.2 & 21.4 & 10.8 \\
 &  & Medium & 60.0 & 27.8 & 0.0 \\
 &  & Hard & 59.1 & 28.9 & 0.0 \\
\cmidrule(l){2-6}
 & \multicolumn{2}{r}{\textit{Overall}} & \textbf{55.3} & \textbf{23.2} & \textbf{8.4} \\
\midrule
\multirow{9}{*}{GPT-5.4} & \multirow{3}{*}{Euclid.} & Easy & 58.3 & 61.5 & 25.0 \\
 &  & Medium & 67.6 & 69.4 & 15.4 \\
 &  & Hard & 61.3 & 70.0 & 20.0 \\
\cmidrule(l){2-6}
 & \multirow{3}{*}{Func.\,\&\,Calc.} & Easy & 62.5 & 3.6 & 12.9 \\
 &  & Medium & 64.3 & 16.7 & 0.0 \\
 &  & Hard & 66.2 & 0.0 & 3.8 \\
\cmidrule(l){2-6}
 & \multirow{3}{*}{Vec.\,\&\,Coord.} & Easy & 51.0 & 30.6 & 34.2 \\
 &  & Medium & 45.2 & 46.5 & 32.4 \\
 &  & Hard & 55.9 & 35.0 & 10.3 \\
\cmidrule(l){2-6}
 & \multicolumn{2}{r}{\textit{Overall}} & \textbf{59.7} & \textbf{40.4} & \textbf{18.2} \\
\midrule
\multirow{9}{*}{GPT-5.4-mini} & \multirow{3}{*}{Euclid.} & Easy & 80.0 & 80.6 & 14.3 \\
 &  & Medium & 69.2 & 80.0 & 0.0 \\
 &  & Hard & 74.2 & 76.5 & 11.1 \\
\cmidrule(l){2-6}
 & \multirow{3}{*}{Func.\,\&\,Calc.} & Easy & 93.9 & 57.1 & 0.0 \\
 &  & Medium & 90.0 & 50.0 & 0.0 \\
 &  & Hard & 97.7 & 87.5 & 0.0 \\
\cmidrule(l){2-6}
 & \multirow{3}{*}{Vec.\,\&\,Coord.} & Easy & 69.4 & 67.6 & 15.4 \\
 &  & Medium & 77.8 & 81.8 & 25.0 \\
 &  & Hard & 80.0 & 80.6 & 0.0 \\
\cmidrule(l){2-6}
 & \multicolumn{2}{r}{\textit{Overall}} & \textbf{83.3} & \textbf{76.3} & \textbf{10.0} \\
\bottomrule
\end{tabular}
\end{table}

\begin{table}[!htbp]
\centering
\small
\setlength{\tabcolsep}{4.2pt}
\caption{
Comparison with closed-source MLLMs, open-source general MLLMs, open-source math-tuned MLLMs, and SFT-only geometry data-generation methods.
All numbers are accuracy (\%). ``--'' indicates that the result or task-specific data scale is not reported under the same benchmark protocol.
Data Scale denotes task-specific math/geometry SFT data when available, rather than generic pretraining data. This table is an extension of Table \ref{tab:downstream_training} of the main paper.
}
\label{tab:downstream_training_ext}
\resizebox{\textwidth}{!}{
\begin{tabular}{lccccc}
\toprule
\textbf{Model} & \textbf{Params} & \textbf{PGPS9K} & \textbf{GeoQA} & \textbf{MathVista-GPS} & \textbf{Data Scale} \\
\midrule
\multicolumn{6}{l}{\textit{Closed-source MLLMs}} \\
Qwen-VL-Plus~\citep{wang2025mathcodervl} & Closed & -- & -- & 35.50 & -- \\
Qwen-VL-Max~\citep{wang2025mathcodervl} & Closed & -- & -- & 46.10 & -- \\
GPT-4V~\citep{deng2025trcot,yang2024mathglm} & Closed & -- & 43.40 & 50.50 & -- \\
Claude-3-Opus~\citep{wang2025mathcodervl} & Closed & -- & -- & 52.90 & -- \\
Gemini Ultra~\citep{deng2025trcot} & Closed & -- & -- & 56.30 & -- \\
GPT-4-Turbo~\citep{wang2025mathcodervl} & Closed & -- & -- & 58.30 & -- \\
Gemini-1.5-Pro~\citep{wang2025mathcodervl} & Closed & -- & -- & 58.90 & -- \\
Claude-3.5-Sonnet~\citep{wang2025mathcodervl} & Closed & -- & -- & 64.40 & -- \\
GPT-4o~\citep{deng2025trcot,wang2025mathcodervl} & Closed & -- & 61.40 & 64.70 & -- \\
\midrule
\multicolumn{6}{l}{\textit{Open-source general MLLMs}} \\
LLaVA-1.5-13B~\citep{wang2025mathcodervl} & 13B & -- & -- & 22.70 & -- \\
DeepSeek-VL-7B~\citep{deng2025trcot,wang2025mathcodervl} & 7B & -- & 33.70 & 28.40 & -- \\
Qwen2-VL-7B/8B~\citep{pan2025geogen,deng2025trcot,wang2025mathcodervl} & 7B/8B & 35.00 & 55.70 & 40.90 & -- \\
InternVL2-8B~\citep{deng2025trcot,wang2025mathcodervl} & 8B & -- & 56.50 & 62.00 & -- \\
InternVL2-26B~\citep{wang2025mathcodervl} & 26B & -- & -- & 54.30 & -- \\
InternVL2-76B~\citep{wang2025mathcodervl} & 76B & -- & -- & 67.80 & -- \\
InternVL2.5-8B~\citep{pan2025geogen,deng2025trcot} & 8B & 37.40 & 59.00 & 67.80 & -- \\
Qwen2.5-VL-7B-Instruct & 7B & 42.40 & 72.07 & 53.37 & -- \\
\midrule
\multicolumn{6}{l}{\textit{Open-source math-tuned MLLMs}} \\
Math-LLaVA-13B~\citep{shi2024mathllava,wang2025mathcodervl} & 13B & -- & 47.80 & 57.70 & 360K \\
MathGLM-Vision-9B~\citep{yang2024mathglm} & 9B & -- & -- & 64.42 & MathVL + VQA \\
MultiMath-7B~\citep{peng2024multimath,wang2025mathcodervl} & 7B & -- & -- & 66.80 & 300K$^\ddagger$ \\
Math-PUMA-Qwen2~\citep{wang2025mathcodervl} & 7B/8B & -- & -- & 48.10 & -- \\
MathCoder-VL-2B~\citep{wang2025mathcodervl} & 2B & -- & -- & 66.40 & 8.6M + 3M \\
MathCoder-VL-8B~\citep{wang2025mathcodervl} & 8B & -- & -- & 73.60 & 8.6M + 3M \\
\midrule
\multicolumn{6}{l}{\textit{SFT-only geometry data-generation methods}} \\
G-LLaVA-7B~\citep{gao2025gllava,deng2025trcot} & 7B & -- & 62.80 & 53.40 & 170K \\
G-LLaVA-13B~\citep{gao2025gllava,pan2025geogen} & 13B & -- & 67.00 & 56.70 & 170K \\
MAVIS-7B~\citep{zhang2025mavis,pan2025geogen} & 7B & -- & 68.30 & 64.10 & 834K \\
TR-CoT-8B~\citep{deng2025trcot,pan2025geogen} & 8B & -- & 75.90 & 73.10 & 87K \\
GeoGen-SFT-3B~\citep{pan2025geogen} & 3B & 44.70 & 75.60 & 64.50 & 224K \\
GeoGen-SFT-7B~\citep{pan2025geogen} & 7B & 54.30 & 78.00 & 74.00 & 224K \\
TR-CoT-Qwen2.5-VL-7B~\citep{deng2025trcot} & 7B & -- & 79.20 & --$^\dagger$ & 65K + Geo170K \\
TR-CoT-InternVL2.5-8B~\citep{deng2025trcot} & 8B & -- & 76.70 & --$^\dagger$ & 65K + Geo170K \\
\midrule
\textbf{VeriGeo-greedy} & 7B & \underline{56.90} & \underline{79.31} & \underline{75.00} & \textbf{8.7k} \\
\textbf{VeriGeo-beam5} & 7B & \textbf{59.40} & \textbf{82.74} & \textbf{75.96} & \textbf{8.7k} \\
\bottomrule
\end{tabular}
}
\vspace{2pt}
\begin{flushleft}
\footnotesize
$^\dagger$ TR-CoT reports MathVista results under its own geometry-problem evaluation setting, but does not explicitly report the exact MathVista-GPS split used in our evaluation; therefore, we do not copy those numbers into the MathVista-GPS column. \\
$^\ddagger$ MultiMath includes a process-supervised RL stage and is listed only as a math-tuned contextual baseline, not as a directly comparable SFT-only geometry data-generation method.\\
$\rhd$ Qwen2.5-VL-7B-Instruct fine-tuned via LoRA ($r=4,\alpha=8$) on 8673 synthetic geometry problems for 1 epoch (205 optimizer steps), AdamW with learning rate $1e-5$, cosine schedule, 200 warmup steps, effective batch size 48, bf16. Inference with beam search (num\_beams=5, top-1).
\end{flushleft}
\end{table}

\clearpage

\begin{figure*}[!htbp]
  \centering
  \includegraphics[width=0.95\textwidth]{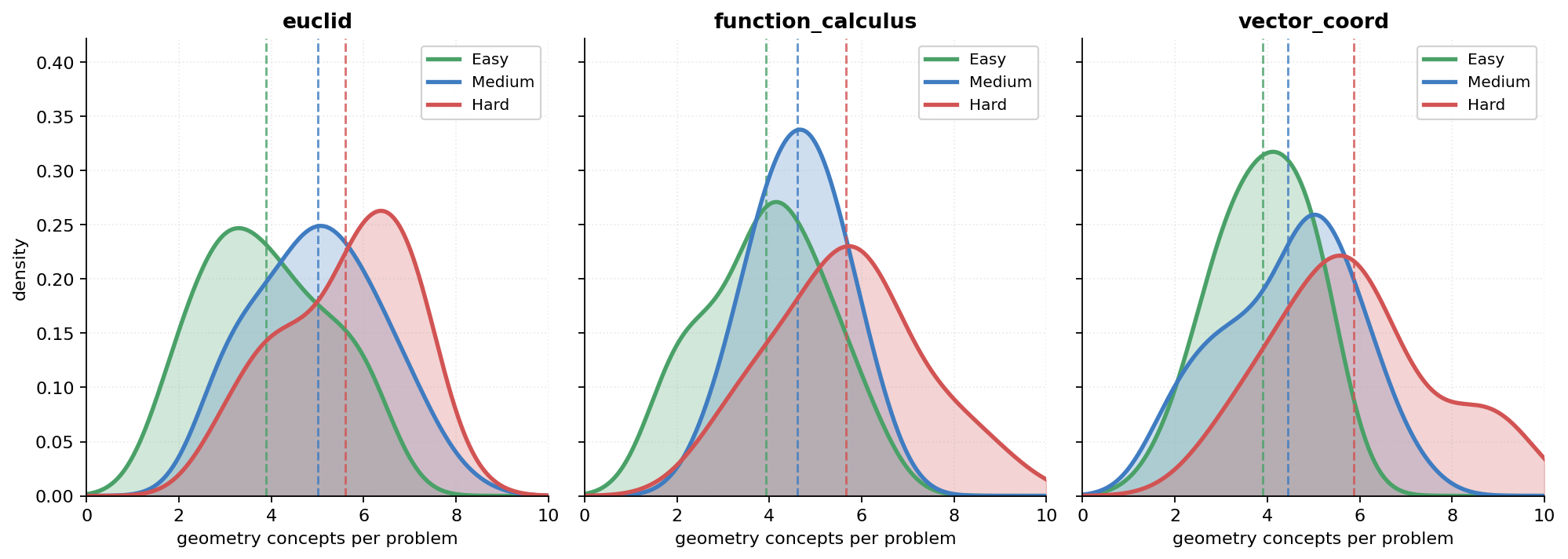}
  \caption{
 Distribution shift of complexity under difficulty control.
  }
  \label{fig:concept_count_drift}
\end{figure*}

\begin{figure}[!htbp]
  \centering
\includegraphics[page=1,width=0.95\textwidth]{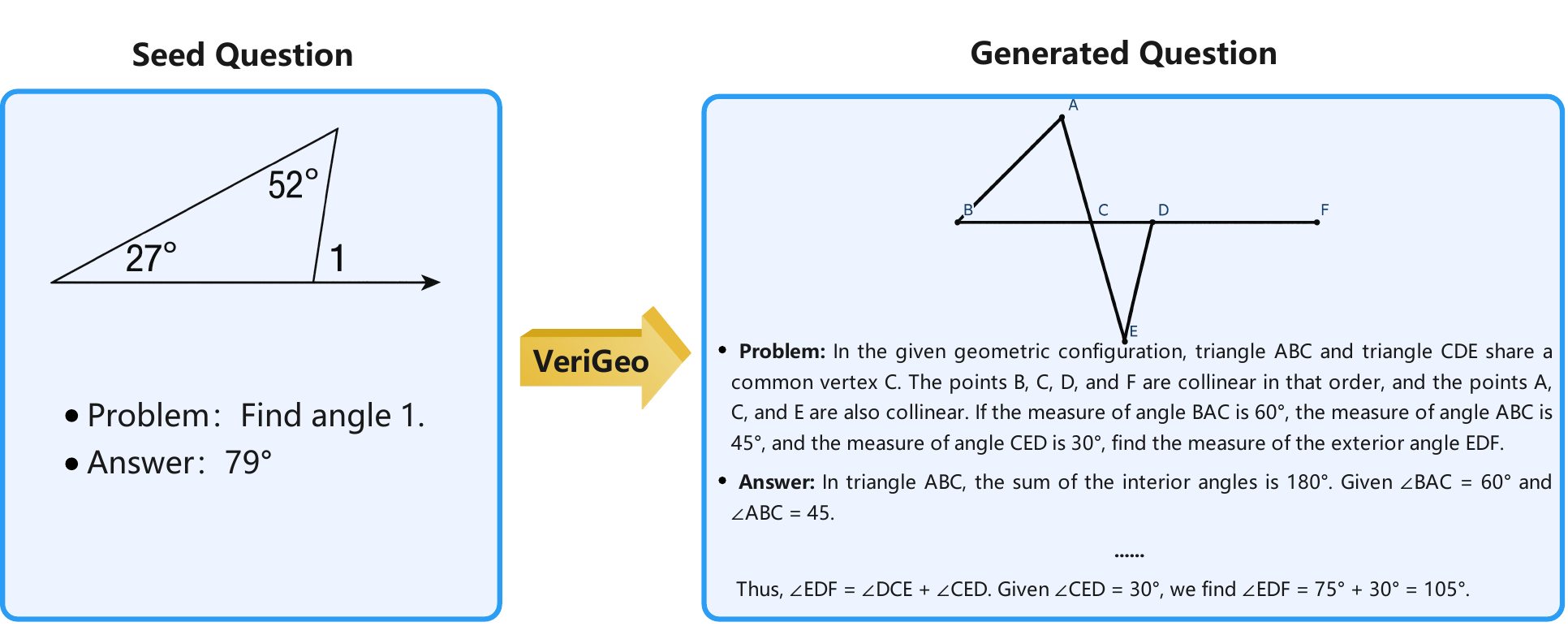}
  \caption{Case study of difficulty modulation: from a one-step triangle angle-sum seed to a multi-constraint angle calculation problem. In this case, the seed problem is a one-step angle calculation task: with two angles in a single triangle given, the target follows directly from the triangle angle-sum rule. In contrast, the difficulty-increased output preserves the same angle calculation theme but embeds it in a richer configuration: two triangles share vertex $C$, with additional collinearity constraints ($B,C,D,F$ and $A,C,E$ are collinear). Solving now requires multiple intermediate inferences. This adds more geometry concepts and a longer reasoning chain while staying aligned with the seed's geometric context.
  }
  \label{fig:case}
\end{figure}

\clearpage


\clearpage
\section{\ourM{} Language Specification}
\label{app:veriscript_spec}
This appendix provides the full grammar, supported operator types, and the stage-specific profiles used by the Author and Solver.

\subsection{Core data model and identifiers}
\ourM{} operates over a shared, mutable geometric state containing typed entities (e.g., \texttt{Point}, \texttt{Segment}, \texttt{Circle}, \texttt{Angle}, \texttt{Triangle}, \texttt{Quadrilateral}, \texttt{Function}) and constraint edges. Each entity is referenced by a string identifier.

\paragraph{Canonical IDs and aliases.}
\ourM{} accepts both canonical IDs and common aliases, then normalizes them before execution. In particular:
\begin{itemize}
  \item Points are atomic IDs (e.g., \texttt{A}, \texttt{P}, \texttt{O}).
  \item Segments are canonicalized as \texttt{SEG\_AB} with endpoints in sorted order; common aliases such as \texttt{AB} are accepted and normalized.
  \item Triangles and quadrilaterals are referenced by explicit registry IDs (e.g., \texttt{TRI\_ABC}, \texttt{RECT\_ABCD}) created via registry operators.
  \item Angles are referenced either by a registered angle ID (e.g., \texttt{ANG\_BAC}) or by a readable alias (e.g., \texttt{$\angle$BAC}) that is normalized to the registered entity.
\end{itemize}

\subsection{Action grammar and JSON envelopes}

\paragraph{Atomic action.}
Every step is a JSON record
\[
a_t \equiv (\texttt{op}, \texttt{type}, \texttt{args}),
\]
where \texttt{op} selects an operator family, \texttt{type} selects a variant, and \texttt{args} supplies ordered arguments (IDs, literals, or expressions). All numerical values are provided as \emph{strings} to ensure determinism across LLM serialization and downstream parsing.

\paragraph{EBNF (trace-level).}
\begin{verbatim}
Action   ::= {"op": Op, "type": Type, "args": [Arg*]}
Arg      ::= string
Actions  ::= [Action*]

AuthorEnvelope ::= {
  "problem": string,
  "givens": [string*],
  "goal": string,
  "answer": string,
  "comment": string,
  "actions": Actions,
  optional "solver_actions": Actions
}

SolverEnvelope ::= {
  "actions": Actions,
  "final_answer": string,
  optional "proof": string
}
\end{verbatim}

\subsection{Stage-specific profiles}
Both stages share the \emph{same} core semantics, but expose different operator subsets in practice.

\paragraph{Author profile (generative).}
Used to \emph{construct} a non-degenerate configuration and a consistent problem package. In addition to introducing entities and constraints, the Author may perform controlled coordinate adjustments (e.g., \texttt{MovePoint}) to improve realizability and avoid degeneracy.

\paragraph{Solver profile (verificative).}
Used to record a \emph{proof-aligned} reasoning trace: derived relations (e.g., \texttt{ExprConstraint}, \texttt{Midpoint}, \texttt{Perpendicular}) plus explicit verification operators (e.g., \texttt{VerifyPoint}, \texttt{VerifyFunction}). The Solver trace is executed stepwise; invalid IDs, missing prerequisites, or failing predicates are rejected immediately.

\subsection{Numerical and expression sublanguage}
Many operator arguments are \texttt{ScalarExpr} strings. The implementation uses a controlled parser that (i) preserves exact intent when possible (e.g., \texttt{"1/2"}, \texttt{"sqrt(2)"}, \texttt{"$\surd$2"}) and (ii) rejects unsafe constructs when evaluating user/LLM-provided function expressions.

\paragraph{ScalarExpr.}
Supported literals include signed integers/decimals, rationals, radicals, and simple arithmetic compositions (e.g., \texttt{"1/2 + sqrt(2)"}). This design avoids float-format ambiguity in JSON and keeps traces reproducible.

\paragraph{Function expressions (for \texttt{AddFunction}/\texttt{VerifyFunction}).}
When a function is specified as \texttt{"y = f(x)"} or as an expression, it is parsed via a restricted AST: only a whitelist of arithmetic operators and elementary functions is permitted; attribute access and other unsafe nodes are disallowed. This makes function-based verification both robust and safe under untrusted generation.

\section{Supplementary Details for the Action Catalog}
\label{app:action_catalog}
A summary of the operations is listed in Table \ref{tab:veriscript_ops}. Details of the operator family, the supported \texttt{type} variants, argument signatures, and effects are summarized below. Unless stated otherwise, all IDs referenced in \texttt{args} must already exist (after normalization). This appendix also includes the (formerly separate) stage-profile comparison, consolidated under the same catalog to avoid redundancy while keeping all original labels intact.

\begin{table}[t]
\centering
\small
\setlength{\tabcolsep}{4pt}
\renewcommand{\arraystretch}{1.05}
\begin{tabularx}{\linewidth}{@{}
  >{\RaggedRight\arraybackslash}p{0.28\linewidth}
  >{\RaggedRight\arraybackslash}X
@{}}
\toprule
\textbf{Family} & \textbf{Representative operators (examples)} \\
\midrule
Point construction &
\texttt{AddPoint}(\texttt{Cartesian}/\texttt{OnSegment}/\texttt{Free}) \\

Primitive geometry &
\texttt{AddAuxLine}(\texttt{Segment}),
\texttt{AddCircle}(\texttt{CenterRadius}/\texttt{CenterPoint}) \\

Object registry &
\texttt{AddAngle}, \texttt{AddTriangle}, \texttt{AddQuadrilateral}(\texttt{Rectangle}/\texttt{Square}/\dots) \\

Constraints &
\texttt{AddEdge}(\texttt{Perpendicular}/\texttt{EqualAngle}/\texttt{ExprConstraint}/\dots) \\

Function modeling &
\texttt{AddFunction}(\texttt{Explicit}),
\texttt{VerifyFunction}(\texttt{DerivativeAt}/\texttt{IntegralOn}) \\

Verification &
\texttt{VerifyPoint}(\texttt{Cartesian}) \\
\bottomrule
\end{tabularx}
\caption{Operator families in \ourM{}}
\label{tab:veriscript_ops}
\end{table}

\paragraph{\texttt{AddPoint} (point construction).}
\begin{itemize}
  \item \textbf{Free}: \texttt{["P"]}. Create a new free point.
  \item \textbf{Cartesian} (Author only): \texttt{["P","x","y"]} with \texttt{x,y: ScalarExpr}. Create a point with explicit coordinates.
  \item \textbf{OnSegment}: \texttt{["P","SEG\_AB","t"]}. Create a point incident to an existing segment; optional seed ratio \texttt{t} (clamped to $[0,1]$) initializes the embedding, while incidence is enforced by constraints.
  \item \textbf{OnCircle}: \texttt{["P","CIR\_O","theta"]}. Create a point on an existing circle; optional seed angle \texttt{theta} (degrees).
  \item \textbf{PointOnCircle}: \texttt{["P","CIR\_O","Q"]}. Create a point on the circle using an existing reference point \texttt{Q} to define a stable initial direction.
\end{itemize}

\paragraph{\texttt{MovePoint} (Author-only coordinate adjustment).}
\textbf{Cartesian}: \texttt{["P","x","y"]}. Update the embedding coordinates of an existing point (used to repair degeneracy or improve layout); constraints remain authoritative and will be re-certified globally.

\paragraph{\texttt{AddAuxLine} (auxiliary segment).}
\textbf{Segment}: \texttt{["A","B"]}. Register (or reuse) the segment between two existing points, creating the canonical ID \texttt{SEG\_AB} after normalization.

\paragraph{\texttt{AddCircle} (primitive geometry).}
\begin{itemize}
  \item \textbf{CenterRadius}: \texttt{["CIR\_O","O","r"]}. Create a circle by center and radius.
  \item \textbf{CenterPoint}: \texttt{["CIR\_O","O","A"]}. Create a circle by center and a point on the circle (radius derived from \texttt{OA}).
\end{itemize}

\paragraph{Registry operators (explicit object creation).}
These operators make composite objects first-class, enabling downstream constraints and expressions to reference them by ID.
\begin{itemize}
  \item \texttt{AddAngle} (\textbf{ByPoints}): \texttt{["ANG","B","A","C"]} registers $\angle BAC$ under ID \texttt{ANG}.
  \item \texttt{AddTriangle} (\textbf{ByPoints}): \texttt{["TRI","A","B","C"]} registers triangle $ABC$ under ID \texttt{TRI}.
  \item \texttt{AddQuadrilateral}: \texttt{["QID","A","B","C","D"]} with one of the following \texttt{type} variants:
    \textbf{Trapezoid}, \textbf{Parallelogram}, \textbf{Rectangle}, \textbf{Square}, \textbf{Rhombus} (and other supported subtypes).
    The registry operator also ensures boundary segments (\texttt{AB}, \texttt{BC}, \texttt{CD}, \texttt{DA}) exist and adds the defining constraints implied by the type.
\end{itemize}

\paragraph{\texttt{AddFunction} (function modeling).}
\textbf{Explicit}: \texttt{["FUNC\_f", "y = f(x)", "x\_min", "x\_max", "samples"]}. Registers an explicit real function (parsed from the right-hand side if \texttt{"y="} is provided) and optional sampling range used for rendering and numerical checks.

\paragraph{\texttt{AddEdge} (constraint edges).}
\texttt{AddEdge} registers a typed constraint edge over an existing scope. The backend supports a broad family of constraints; in \emph{strict} mode, a selected subset is compiled into equation systems (Appendix~\ref{app:verisolve_equations}), while the remaining relations are validated by certified numerical checks and semantic passes (Appendices~\ref{app:verisolve_numeric} and \ref{app:verisolve_semantic}).
Common \texttt{type} variants include:
\begin{itemize}
  \item \textbf{Segment relations} (\texttt{SEG\_..} or point-pair forms are accepted and normalized):
  \texttt{Perpendicular}, \texttt{Parallel}, \texttt{EqualLength}.
  Signature: \texttt{[seg1, seg2]}.
  \item \textbf{Angle relations}:
  \texttt{EqualAngle}. Signature: \texttt{[ang1, ang2]}.
  \item \textbf{Incidence / collinearity / locus}:
  \texttt{Collinear} (\texttt{[P1,P2,P3,...]}),
  \texttt{PointOnLine} (\texttt{[P, SEG\_AB]}),
  \texttt{PointOnSegment} (\texttt{[P, SEG\_AB, t]}),
  \texttt{OnCircle} (\texttt{[P, CIR\_O]}).
  Internally, these are normalized into a uniform incidence representation with optional parameters (e.g., segment ratio).
  \item \textbf{Midpoint / angle bisector}:
  \texttt{Midpoint} (\texttt{[M, SEG\_AB]} or \texttt{[M, A, B]}),
  \texttt{AngleBisector} (\texttt{[SEG\_AE, ANG\_BAC]} and common shorthands that normalize to this form).
  \item \textbf{Cyclic / concyclic}:
  \texttt{Cyclic} (\texttt{[A,B,C,D]}) is supported as an alias that is normalized to concyclicity checks and/or circle-incidence constraints depending on whether an explicit circle object exists.
  \item \textbf{Triangle relations}:
  \texttt{SimilarTriangles}, \texttt{CongruentTriangles}. Signature: \texttt{[TRI\_ABC, TRI\_DEF]}.
  \item \textbf{Vector constraints} (coordinate geometry):
  \begin{itemize}
    \item \texttt{VectorSum}: \texttt{[A,B, C,D, E,F]} encoding $\overrightarrow{AB}=\overrightarrow{CD}+\overrightarrow{EF}$.
    \item \texttt{VectorDiff}: \texttt{[A,B, C,D, E,F]} encoding $\overrightarrow{AB}=\overrightarrow{CD}-\overrightarrow{EF}$.
    \item \texttt{VectorDotZero}: \texttt{[A,B, C,D]} encoding $\overrightarrow{AB}\cdot\overrightarrow{CD}=0$.
    \item \texttt{VectorLinear}: \texttt{[A,B, k1,C1,D1, k2,C2,D2, ...]} encoding
    $\overrightarrow{AB}=\sum_i k_i\,\overrightarrow{C_iD_i}$.
  \end{itemize}
  \item \textbf{Algebraic constraints}:
  \texttt{ExprConstraint}. Signature: \texttt{[lhs, rel, rhs, tol]}, where
  \texttt{rel} is one of \texttt{=, ==, !=, <=, >=, <, >} and \texttt{tol} is an optional \texttt{ScalarExpr}.
  \item \textbf{Advanced relations} (validated by certified checks when present):
  \texttt{Tangent}, \texttt{PowerEquality}, and other higher-level theorem-backed relations used by the Solver trace and/or derived-constraint closure.
\end{itemize}

\paragraph{\texttt{VerifyPoint} (explicit coordinate check; Solver profile).}
\textbf{Cartesian}: \texttt{["P","x","y","tol"]}. Checks that the realized point coordinates match the claimed values within tolerance.

\paragraph{\texttt{VerifyFunction} (function reasoning; Solver profile).}
\begin{itemize}
  \item \textbf{PointsOnFunction}: \texttt{["f", P1, P2, ..., "tol"]} where each \texttt{Pi} may be a point ID or a coordinate pair \texttt{"(x,y)"}; verifies pointwise satisfaction of $y=f(x)$.
  \item \textbf{DerivativeAt}: \texttt{["f","x0","df","n=order","tol"]}. Numerically checks an $n$-th derivative at \texttt{x0} (or at a point ID via its \texttt{x}-coordinate).
  \item \textbf{IntegralOn}: \texttt{["f","a","b","val","tol"]}. Numerically checks $\int_a^b f(x)\,dx$; bounds may be numeric, point IDs (using their \texttt{x}-coordinate), or infinities (\texttt{"inf"}, \texttt{"-inf"}), with optional breakpoints for piecewise evaluation.
\end{itemize}

\subsection{Comparison of Actions/Operators  used in Author and Solver Agents}
\label{app:verisolve_ops}

\paragraph{Stage-specific profiles in code (operator-level separation).}
The implementation enforces \emph{distinct allowed operator sets} for Author vs.\ Solver traces, while keeping a shared execution core:
\begin{itemize}
  \item \textbf{Author (generative) trace} permits diagram construction and controlled layout repair:
    \[
    \begin{aligned}
    \texttt{AllowedOps}_{\textsf{Author}} = \{&
    \texttt{AddPoint, MovePoint, AddAuxLine, AddEdge,AddCircle,} \\
    &\texttt{ AddAngle, AddTriangle,AddQuadrilateral, AddFunction}
    \}.
    \end{aligned}
    \]

  \item \textbf{Solver (verificative) trace} is restricted to proof-aligned, checkable steps:
  \[
  \texttt{AllowedOps}_{\textsf{Solver}}=\{\texttt{AddEdge, AddPoint, AddAuxLine, AddFunction, VerifyPoint, VerifyFunction}\}.
  \]
\end{itemize}
This separation is \emph{enforced twice}: (i) at LLM generation time via strict JSON-schema response formats, and (ii) at runtime by a validator that rejects out-of-profile operators, malformed signatures, or unresolved IDs.

\paragraph{Normalization and robustness under LLM noise.}
To make traces stable under minor formatting variation, the parser performs conservative normalizations without changing semantics:
\begin{itemize}
  \item Canonicalizes segment/angle IDs (\texttt{AB} $\rightarrow$ \texttt{SEG\_AB}; readable angle tokens $\rightarrow$ registered vertex triplets).
  \item Sanitizes argument typing: numerical fields are accepted as strings/ints/floats but are stored and processed as strings, ensuring deterministic downstream parsing.
  \item Repairs common shorthands (e.g., \texttt{AddAuxLine} that redundantly includes a segment name in \texttt{args}) and then revalidates the normalized action.
\end{itemize}
Crucially, these repairs are \emph{not} heuristic acceptance of invalid steps: if a normalization cannot map the action into a valid canonical form, the step is rejected.

\paragraph{Proof--action alignment as a verifiability constraint.}
The Solver prompt and validator jointly enforce that every non-trivial numerical or algebraic conclusion in the proof must be emitted as an explicit \texttt{ExprConstraint} step. This makes the reasoning trace executable, auditable, and \emph{replayable} (rather than a free-form narrative), which is essential for reliable LLM supervision and for reviewer-facing inspection.

\paragraph{Conservative derived-constraint closure (optional but impactful).}
Beyond user-provided constraints, the backend optionally applies a provenance-tagged closure pass (\texttt{source=derived}) that adds \emph{safe} implied relations when prerequisites are already certified---e.g., inferring \texttt{SimilarTriangles} from two parallel-induced angle correspondences, or from numerically certified AA matches. Because every derived edge is guarded by existence checks and tagged with its reason, this augmentation strengthens downstream verification without introducing silent assumptions.

\paragraph{Minimal examples (illustrative only).}
\begin{quote}\small
\texttt{\{"op":"AddPoint","type":"Cartesian","args":["A","0","0"]\}} \hfill (\textsf{Author})\\
\texttt{\{"op":"AddEdge","type":"Perpendicular","args":["SEG\_AB","SEG\_CD"]\}} \hfill (\textsf{Both})\\
\texttt{\{"op":"VerifyFunction","type":"DerivativeAt","args":["FUNC\_f","x=1","2","n=1","1e-4"]\}} \hfill (\textsf{Solver})\\
\texttt{\{"op":"AddQuadrilateral","type":"Square","args":["SQ\_ABCD","A","B","C","D"]\}} \hfill (\textsf{Author})
\end{quote}

\subsection{Human-Readable Exact Quantities}\label{sec:digital_num}
A persistent challenge in LLM-based geometry synthesis is the instability of float numbers: a pipeline may be logically correct yet fail verification because calculation artifacts perturb downstream checks or yield non-canonical answers. To address this, we canonicalize numerical quantities into human-readable, verifiable forms: (i) integers/decimals (e.g., \texttt{"3"}, \texttt{"0.5"}); (ii) rationals (e.g., \texttt{"-2/3"}); (iii) radicals (e.g., \texttt{"\textbackslash{}sqrt(2)"}); and (iv) simple arithmetic over constants (e.g., \texttt{"1/2 + sqrt(2)"}). This approach mitigates precision errors introduced by floating-point arithmetic and ensures the output is more user-friendly.


\section{Supplementary Information of Analytical Verification}
\subsection{Analytical Constraints}
\label{app:verisolve_equations}

Let each point $P$ have unknowns $(x_P,y_P)$ unless already \texttt{known} or closed-form locked.
Our analytical solver compiles the following constraint types into equations:

\paragraph{Collinear.}
For each triple $(A,B,C)$ in scope:
\[
(x_B-x_A)(y_C-y_A)-(y_B-y_A)(x_C-x_A)=0.
\]

\paragraph{Parallel.}
For segments $\overline{AB}$ and $\overline{CD}$:
\[
(x_B-x_A)(y_D-y_C)-(y_B-y_A)(x_D-x_C)=0.
\]

\paragraph{Perpendicular.}
\[
(x_B-x_A)(x_D-x_C)+(y_B-y_A)(y_D-y_C)=0.
\]

\paragraph{EqualLength.}
\[
\|B-A\|^2-\|D-C\|^2=0.
\]

\paragraph{EqualAngle.}
Given angle tokens resolvable into vertex triplets $(A,B,C)$ and $(D,E,F)$,
the engine attempts the signed-cosine form:
\[
\frac{(A-B)\cdot(C-B)}{\|A-B\|\|C-B\|}
-
\frac{(D-E)\cdot(F-E)}{\|D-E\|\|F-E\|}=0,
\]
and otherwise falls back to a polynomial equality derived from dot products and squared lengths:
\[
\bigl((A-B)\cdot(C-B)\bigr)^2 \|D-E\|^2\|F-E\|^2
-
\bigl((D-E)\cdot(F-E)\bigr)^2 \|A-B\|^2\|C-B\|^2=0.
\]

\paragraph{AngleBisector.}
For an angle token $(X,Y,Z)$ and a bisector segment that shares vertex $Y$,
let the other endpoint be $E$; the engine enforces:
\[
\cos\angle XYE-\cos\angle EYZ=0
\]
(using the same cosine fallback rules as \textsf{EqualAngle}).

\paragraph{VectorDotZero.}
For $\overrightarrow{AB}\cdot\overrightarrow{CD}=0$:
\[
(x_B-x_A)(x_D-x_C)+(y_B-y_A)(y_D-y_C)=0.
\]

\paragraph{Midpoint.}
For midpoint $M$ of $AB$:
\[
2x_M-x_A-x_B=0,\quad 2y_M-y_A-y_B=0.
\]

\paragraph{PointOnLine.}
For point $P$ on line/segment with reference endpoints $A,B$:
\[
(x_P-x_A)(y_B-y_A)-(y_P-y_A)(x_B-x_A)=0.
\]

\paragraph{PointOnSegment (ratio-locked or free).}
If ratio $r$ is provided:
\[
x_P-x_A-r(x_B-x_A)=0,\quad y_P-y_A-r(y_B-y_A)=0.
\]
Otherwise, a free scalar parameter $t$ is introduced and optimized jointly
with other variables under the same affine constraints.

\paragraph{Incidence.}
If target is a circle, it reduces to \textsf{OnCircle}.
If target is a line/segment, it reduces to \textsf{PointOnLine}.

\paragraph{OnCircle / PointOnCircle.}
Let circle center be $O$; the engine uses $r^2$ as:
(i) a constant if numerical radius exists; else (ii) $\|Q-O\|^2$ if a through-point $Q$ exists; else
(iii) a fresh symbolic parameter.
Then it enforces $(x_P-x_O)^2+(y_P-y_O)^2-r^2=0$.

\paragraph{Concyclic.}
If the scope is of the form $(P,\mathsf{CIR})$ where $\mathsf{CIR}$ is an explicit circle entity,
the constraint is reduced to \textsf{OnCircle}.
Otherwise, let the in-scope point list be $(P_1,\dots,P_m)$ (non-point tokens are ignored).
When $m\ge 4$, the engine fixes the first three points $(A,B,C)=(P_1,P_2,P_3)$ as a reference
and for each remaining point $D\in\{P_4,\dots,P_m\}$ enforces the determinant form:
\[
\det
\begin{pmatrix}
x_A^2+y_A^2 & x_A & y_A & 1\\
x_B^2+y_B^2 & x_B & y_B & 1\\
x_C^2+y_C^2 & x_C & y_C & 1\\
x_D^2+y_D^2 & x_D & y_D & 1
\end{pmatrix}
=0,
\qquad \forall\, D\in\{P_4,\dots,P_m\}.
\]
(Equivalently, the analytical engine appends $\det(M_{A,B,C,D})$ as a polynomial equation for each extra point.)

\paragraph{Gauge constraints.}
If fewer than two anchored points exist, the engine fixes an origin and scale:
either $(x_{P_0},y_{P_0})=(0,0)$ and $(x_{P_1},y_{P_1})=(1,0)$, or if one anchor exists,
it pins a second point to be one unit to the right on the same horizontal line.

\paragraph{Closed-form pre-locking (before global solving).}
Before materializing the full symbolic system, our analytical solver opportunistically locks a small set of
\emph{single-unknown} patterns in closed form to reduce search space and avoid degenerate roots.
A representative case is a \emph{translation-locked} segment: if a reference segment $\overline{AB}$
has both endpoints anchored and a target segment has one anchored endpoint $C$ with the other endpoint $D$ unknown,
the solver may directly set
\[
(x_D,y_D)=(x_C,y_C) + \bigl((x_B-x_A),(y_B-y_A)\bigr),
\]
provided that $D$ is not simultaneously entangled by multiple unrelated constraints.
To keep the verification uniform, such locked results are injected as equality equations (rather than
directly writing coordinates) whenever the point is not yet hard-locked.

\paragraph{Equation sanitization.}
All constraints are converted into a zero-equality form $f(\mathbf{z})=0$.
If an equation becomes constant (no free symbols), the solver treats it as an immediate consistency check:
it is discarded when numerically satisfied, and the instance is rejected when violated.

\paragraph{Hybrid solve with timeout and numerical fallback.}
Our analytical solver first attempts a symbolic solve under a strict time budget.
If the symbolic attempt fails, times out, or yields a degenerate assignment that cannot pass residual checks,
it falls back to a numerical root finder (e.g., \texttt{nsolve}) with randomized initializations and
accepts the first solution that satisfies all constraints within tolerance.

\paragraph{Underdetermined completion (free-parameter handling).}
When the system is underdetermined (fewer independent equations than unknowns),
our analytical solver introduces auxiliary parameters (e.g., for \textsf{PointOnSegment} or unknown $r^2$ in circles)
and assigns a deterministic fill constant to a small subset of low-frequency free variables to obtain
a concrete embedding. Additionally, for a perpendicular constraint where the shared endpoint is still free,
a geometric completion may be applied by placing the point at a Thales construction over the two known endpoints,
i.e.,
\[
U = M \pm \frac{\|A-B\|}{2}\, \mathbf{n}, \quad M=\frac{A+B}{2},
\]
with a fixed sign convention to avoid point collapse.

\paragraph{Residual check and safe commit.}
A candidate solution is accepted only if every equation evaluates to a scalar residual below a global threshold.
Upon acceptance, the analytical solver commits each newly solved point \emph{once} and respects protected points unless
explicitly overridden; committed points are added to a hard-lock list for downstream stages.

\paragraph{Circle synchronization and second-intersection repair.}
After committing coordinates, circle radii are synchronized from the current center/through-point definition.
To address the common ambiguity of circle--line intersections (two valid roots), our analytical solver includes a post-pass that
detects points stuck at the anchor/root under \textsf{OnCircle}+\textsf{Collinear}-style constraints and
moves them to the alternative intersection whenever required for consistency.

\subsection{Rank-aware solving and underdetermined completion}
\label{app:verisolve_solve}

\paragraph{Component splitting.}
Given equation list $\mathcal{E}$ over symbols $\mathcal{S}$, the analytical solver builds a bipartite graph between equations
and the symbols appearing in them, and solves each connected component independently. This reduces blow-up and allows
local fallbacks.

\paragraph{Rank estimation and independent equation selection.}
For a component, the engine estimates rank by numerically evaluating the Jacobian $J$ at random assignments
and computing $\mathrm{rank}(J)$.
It then greedily selects a subset of equations that increases rank to (approximately) the target rank,
improving robustness under redundancy.

\paragraph{Solvers and acceptance.}
The engine first calls symbolic \texttt{solve}. If it fails, it falls back to \texttt{nsolve} with randomized
initializations and accepts a solution iff the residual of \emph{all} compiled equations is below \texttt{analytic\_tol}.
If a component remains underdetermined, remaining free symbols are filled deterministically with
\texttt{analytic\_fill\_constant} (default $0.35$), optionally attempting partial solves first
(\texttt{analytic\_fill\_free\_params=True}).

\subsection{Closed-form warm starts and repairs}
\label{app:verisolve_closedform}

Our analytical solver applies closed-form placements before global solving:
\begin{itemize}
  \item \textbf{Midpoint:} $M=\frac{A+B}{2}$ when both endpoints are numeric.
  \item \textbf{Ratio-locked on segment:} $P=A+r(B-A)$ when $r$ is provided.
  \item \textbf{Triangle canonical points (if $A,B,C$ are known):}
  \begin{itemize}
    \item Incenter $I$ via side-length barycentric weights:
    $I=\frac{aA+bB+cC}{a+b+c}$ with $a=\|B-C\|$, $b=\|A-C\|$, $c=\|A-B\|$.
    \item A specific bisector--circumcircle second intersection $E$ computed by intersecting the
    $A$-angle-bisector ray with the circumcircle and choosing the non-$A$ intersection.
    \item Thales-style points $N,P$ on the circle with diameter $EB$ and $EC$, respectively, with a fixed sign rule
    to avoid coincident flips.
  \end{itemize}
  \item \textbf{Circle--line second intersection repair:}
  if a point is constrained by \textsf{Concyclic} and \textsf{Collinear} and collapses to the anchor intersection,
  choose the farther non-anchor intersection on the circle.
  \item \textbf{Parallel single-unknown completion:}
  when a parallel constraint involves a segment with one unknown endpoint that participates in no other constraints,
  translate the known endpoint by the reference direction vector.
\end{itemize}

\subsection{Non-strict mode: incidence completion and semantic intersection passes}
\label{app:verisolve_semantic}

\paragraph{Intersection resolution.}
Given a point $P$ constrained by intersections of
(1) line--line, (2) circle--line, or (3) circle--circle,
Our analytical solver enumerates candidate intersection points and chooses:
(i) a candidate within segment bounds if applicable,
(ii) a candidate not already used by another point (via \texttt{used\_targets}),
and (iii) the candidate closest to $P$'s previous coordinates to stabilize layouts.
A minimum separation threshold avoids multiple points collapsing to the same location.

\paragraph{Semantic passes (implemented as four passes).}
After raw intersections, the analytical solver performs semantic placements for common constructions:
\begin{itemize}
  \item \textbf{Pass 1 (local endpoints / midpoints):} ensure auxiliary points tied to a segment are consistent with its
  current endpoints and midpoint definitions.
  \item \textbf{Pass 2 (triangle foot / bisector intersection):} compute foot point $H$ and bisector intersection $L$
  from available $(A,B,C)$-like configurations using projection / line intersection.
  \item \textbf{Pass 3 (circumcircle and tangent proxies):} compute a designated second intersection $P$ of a bisector
  with circumcircle; compute $S$ as the intersection of a tangent-at-$A$ direction with $BC$; and set point $R$ on the
  tangent line through $A$.
  \item \textbf{Pass 4 (direction correction):} refine cached directions for segments in \textsf{Parallel}/\textsf{Perpendicular}
  relations to reduce accumulated drift.
\end{itemize}
All semantic writes respect \texttt{hard\_locked} points and do not overwrite protected points unless explicitly allowed.

\section{Numerical verification library}
\label{app:verisolve_numeric}

Numerical verification provides \emph{deterministic, per-constraint} residual checks over the embedded diagram produced by
our analytical solver (and its repair loops). Concretely, \texttt{verify\_constraints(ang\_tol\_deg, len\_tol, circle\_tol)} returns a
dictionary keyed by constraint ID, where each entry contains (i) a three-valued verdict
$\texttt{ok}\in\{\texttt{True},\texttt{False},\texttt{None}\}$, (ii) a diagnostic residual string \texttt{msg}, and (iii) the
constraint \texttt{type}, \texttt{scope}, and \texttt{source}. We intentionally use \texttt{ok=None} for unsupported or
unchecked constraint types to avoid falsely asserting correctness when a numerical predicate is not implemented.

\paragraph{Shared geometric primitives.}
Let $P=(x_P,y_P)$ denote a point coordinate, $\|u\|_2$ the Euclidean norm, and
$\overrightarrow{PQ}=(x_Q-x_P,\,y_Q-y_P)$.
We use:
(i) distances $d(P,Q)=\|P-Q\|_2$;
(ii) angles $\angle(u,v)=\arccos\!\big(\frac{u\cdot v}{\|u\|_2\|v\|_2}\big)$ in degrees, with degeneracy checks when
$\|u\|_2$ or $\|v\|_2$ is near zero;
(iii) segment endpoints inferred from \textsf{Incidence} constraints (and token fallbacks for \texttt{SEG\_AB});
(iv) point--line distance
$\mathrm{dist}(P,\overline{AB})=\frac{|(P-A)\times(B-A)|}{\|B-A\|_2}$ and its projection parameter
$t=\frac{(P-A)\cdot(B-A)}{\|B-A\|_2^2}$ (used for segment-bounded checks).

\paragraph{Implemented numerical checks.}
All checks are scale-aware via tolerances \texttt{ang\_tol\_deg}, \texttt{len\_tol}, and \texttt{circle\_tol}, and each check
reports a residual magnitude to support targeted repair and auditing.

\paragraph{Circle incidence.}
For \textsf{OnCircle}$(P,\mathrm{CIR}_O)$ (and the 2-argument \textsf{Concyclic} encoding used in our engine), we verify
\[
\big|d(P,O)-r\big| \le \texttt{circle\_tol},
\]
where $(O,r)$ are retrieved from the circle entity. For legacy multi-point concyclicity scopes (3/4 points), we conservatively
mark the constraint as skipped (compatibility-safe) rather than raising an exception.

\paragraph{Metric, angular, and midpoint constraints.}
\begin{itemize}
  \item \textsf{EqualLength}$(\overline{AB},\overline{CD})$: verify $|d(A,B)-d(C,D)|\le \texttt{len\_tol}$.
  \item \textsf{Parallel}$(\overline{AB},\overline{CD})$: let $\theta=\angle(\overrightarrow{AB},\overrightarrow{CD})$ and
  check $\min(\theta,|180^\circ-\theta|)\le \texttt{ang\_tol\_deg}$.
  \item \textsf{Perpendicular}$(\overline{AB},\overline{CD})$: check $|90^\circ-\angle(\overrightarrow{AB},\overrightarrow{CD})|
  \le \texttt{ang\_tol\_deg}$.
  \item \textsf{EqualAngle}$(\angle XYZ,\angle DEF)$: parse each angle token into vertex triplets and verify
  $|\angle XYZ-\angle DEF|\le \texttt{ang\_tol\_deg}$.
  \item \textsf{Midpoint}$(M,\overline{UV})$: verify $|d(M,U)-d(M,V)|\le \texttt{len\_tol}$.
  \item \textsf{AngleBisector}$(\overline{YE},\angle XYZ)$: using the named angle entity (e.g., ``$\angle XYZ$''),
  verify $|\angle XYE-\angle EYZ|\le \texttt{ang\_tol\_deg}$.
\end{itemize}

\paragraph{Collinearity and incidence.}
\begin{itemize}
  \item \textsf{Collinear}$(A,B,C)$: compute twice the signed area
  $A_2=\big|(B-A)\times(C-A)\big|$ and check
  \[
    A_2 \le \texttt{len\_tol}\cdot \max\!\big(d(A,B),d(B,C),d(C,A),1\big),
  \]
  which is more scale-stable than a fixed absolute threshold.
  \item \textsf{PointOnLine}$(P,\overleftrightarrow{AB})$: check $\mathrm{dist}(P,\overleftrightarrow{AB})\le \texttt{len\_tol}$.
  \item \textsf{PointOnSegment}$(P,\overline{AB})$: check both $\mathrm{dist}(P,\overleftrightarrow{AB})\le \texttt{len\_tol}$
  and $t\in[-10^{-3},\,1+10^{-3}]$ for the projection parameter $t$ to enforce boundedness with a small numerical margin.
\end{itemize}

\paragraph{Vector relations.}
Let $\overrightarrow{PQ}$ be defined from point coordinates. We use $\|\Delta v\|_2$ as the residual for all vector equalities.
\begin{itemize}
  \item \textsf{VectorSum}: $\overrightarrow{UV}=\overrightarrow{PQ}+\overrightarrow{RS}$, check
  $\|\overrightarrow{UV}-(\overrightarrow{PQ}+\overrightarrow{RS})\|_2\le \texttt{len\_tol}$.
  \item \textsf{VectorDiff}: $\overrightarrow{UV}=\overrightarrow{PQ}-\overrightarrow{RS}$, check
  $\|\overrightarrow{UV}-(\overrightarrow{PQ}-\overrightarrow{RS})\|_2\le \texttt{len\_tol}$.
  \item \textsf{VectorDotZero}: $\overrightarrow{PQ}\cdot\overrightarrow{RS}=0$, check $|\overrightarrow{PQ}\cdot\overrightarrow{RS}|
  \le \texttt{len\_tol}$.
  \item \textsf{VectorLinear}: $\overrightarrow{UV}=\sum_i k_i\,\overrightarrow{P_iQ_i}$ (coefficients parsed as floats),
  check $\big\|\overrightarrow{UV}-\sum_i k_i\overrightarrow{P_iQ_i}\big\|_2\le \texttt{len\_tol}$.
\end{itemize}

\paragraph{Triangle relations.}
For $\triangle ABC$ and $\triangle DEF$, let side-length vectors be
$\ell_1=(|AB|,|BC|,|CA|)$ and $\ell_2=(|DE|,|EF|,|FD|)$, and angle vectors be
$\alpha_1=(\angle A,\angle B,\angle C)$ and $\alpha_2=(\angle D,\angle E,\angle F)$, computed from coordinates.
Degenerate cases (near-zero sides/undefined angles) are rejected.
\begin{itemize}
  \item \textsf{SimilarTriangles}: compute ratios $r_i=\ell_{1,i}/\ell_{2,i}$, set $\bar r=\frac{1}{3}\sum_i r_i$, and check
  $\max_i |r_i-\bar r|/\max(|\bar r|,10^{-6}) \le 5\times 10^{-3}$ and
  $\max_i|\alpha_{1,i}-\alpha_{2,i}|\le \texttt{ang\_tol\_deg}$.
  \item \textsf{CongruentTriangles}: check $\max_i|\ell_{1,i}-\ell_{2,i}|\le \texttt{len\_tol}$.
\end{itemize}

\paragraph{Circle theorems.}
\begin{itemize}
  \item \textsf{Tangent}$(\overline{t},\mathrm{CIR}_O,T)$: verify (i) $T$ lies on the circle within \texttt{circle\_tol}, and
  (ii) the tangent direction at $T$ is perpendicular to the radius $\overrightarrow{OT}$:
  $|90^\circ-\angle(\overrightarrow{OT},\overrightarrow{TT'})|\le \texttt{ang\_tol\_deg}$, where $T'$ is the farthest other
  incidence point on segment $\overline{t}$.
  \item \textsf{PowerEquality}$(P,\mathrm{CIR}_O,\overline{s},\overline{t})$: let $U,V$ be the two circle points incident to the
  secant segment $\overline{s}$ (chosen closest to $P$), and let $T$ be a circle point incident to the tangent segment
  $\overline{t}$ that also contains $P$. We verify
  \[
    \big|\,|PU|\cdot|PV| - |PT|^2\,\big| \le \tau,
    \qquad
    \tau=\max\big(\texttt{len\_tol}\cdot \max(|PU|\cdot|PV|,|PT|^2,1),\ \texttt{len\_tol}\big),
  \]
  i.e., an error budget that scales with the magnitude of the quantities being compared.
\end{itemize}

\paragraph{Expression constraints (\textsf{ExprConstraint}).}
Beyond fixed geometric predicates, we support a unified numerical checker for declarative expressions of scalars, 2D vectors,
and complex-valued vectors. Given \textsf{ExprConstraint} metadata $(\texttt{lhs},\texttt{relation},\texttt{rhs},\texttt{tolerance})$,
we evaluate both sides using an expression evaluator (e.g., \texttt{Length()}, \texttt{Angle()}, \texttt{Area()},
vector/complex utilities) and then check:
(i) vector equality/inequality using $\|\Delta v\|_2$ with a \emph{scale-adaptive} default tolerance
$\max(10^{-6},10^{-6}\cdot\max(\|v_\texttt{lhs}\|_2,\|v_\texttt{rhs}\|_2,1))$;
(ii) complex equality/inequality using $|z_\texttt{lhs}-z_\texttt{rhs}|$ with the same scale rule;
(iii) scalar comparisons over $\{=,\neq,<,\le,>,\ge\}$ with an additive tolerance margin.
To improve robustness in mixed textual/diagram inputs, angle tokens such as ``$\angle XYZ$'' are normalized into a canonical
form before evaluation, and arithmetic relations written as operator forms (e.g., ``\texttt{lhs + something = rhs}'') are
rewritten into standard comparisons prior to checking. Each entry records the evaluated values and the effective tolerance,
enabling reproducible debugging.

\paragraph{Hybrid symbolic rescue for floating-point stability.}
For a subset of geometry primitives
\{\textsf{EqualLength}, \textsf{Parallel}, \textsf{Perpendicular}, \textsf{Collinear}, \textsf{PointOnLine}, \textsf{PointOnSegment}\},
we additionally implement a \emph{symbolic rescue} path: when the numerical check returns \texttt{False}, we convert point
coordinates into rational numbers via \texttt{Fraction(str(x)).limit\_denominator(1024)} (with caching) and re-evaluate the
predicate exactly using dot/cross products and rational arithmetic. This hybrid design materially reduces false negatives
from floating-point noise while preserving the speed and diagnostic richness of the primary numerical verifier.


\section{Calibration Exemplars for the Few-Shot Difficulty Judge}
\label{app:calibration-shots}

The few-shot difficulty judge described in Section~\ref{sec:Blueprint Generation} is
parameterised by nine exemplar problems---three per tier
(\textsc{Easy}, \textsc{Medium}, \textsc{Hard})---which jointly define
the classifier's notion of difficulty. To verify that the judge is
not overly sensitive to any particular choice of exemplars, we
manually constructed three calibration sets and report their
target-tier matching rates in Table~\ref{tab:qwen35plus_difficulty_match}. 
Set~1 uses the same exemplars as those used for question generation, while Sets~2 and~3 are independently constructed exemplar sets used only for robustness evaluation. The three sets are pairwise disjoint (27 distinct problems in total); all
items were written by the authors and cross-checked by a second
reviewer before inclusion.

\paragraph{Selection criteria.} For each set, we selected items that
collectively cover the three problem families studied in VeriGeo
(\textsc{Euclidean}, \textsc{Function~\&~Calculus}, and
\textsc{Vector~\&~Coordinate}), and required every item to satisfy
the reasoning-depth convention of its tier: \textsc{Easy} problems
resolve with a single named theorem or a direct substitution;
\textsc{Medium} problems chain two to three theorems with at most one
derived point; \textsc{Hard} problems chain four or more reasoning
steps, typically over multiple derived points or a non-trivial
geometric invariant. Two authors independently scored each candidate
on these criteria, and disagreements were reconciled through
discussion before the item was admitted into a set.

Set 1 is the examples we used for question geenration. Set 2 and Set 3 are two additional sets used to verify the robuness that we choose another two sets to see whether our questions verifeid are stable. 

\subsection{Set 1}
\label{app:shots:set1}
Set~1 serves a dual role: it provides the difficulty exemplars used to guide problem generation and also calibrates the few-shot difficulty judge for evaluation.

\paragraph{Easy~/~Euclidean.}
\textit{Problem.} In triangle $ABC$, $D,E$ lie on $AB,AC$ with $AD=3$,
$DB=6$, $AE=4$, $EC=8$; $\angle ABC = 50^\circ$, $\angle ACB = 70^\circ$.
Find $\angle DEC$.\\
\textit{Why selected.} The proportion $AD/AB = AE/AC$ gives
$DE \parallel BC$ in one step; the answer follows from a single
parallel-line angle relation---a canonical one-step \textsc{Easy}.

\paragraph{Easy~/~Function~\&~Calculus.}
\textit{Problem.} $f(x) = x^2 - 4$. Points $A, B$ lie on the curve at
$x=1, 3$; vertical drops from $A, B$ hit the $x$-axis at $C, D$.
Segment $AB$ connects $A, B$.\\
\textit{Why selected.} Every length, slope, and area reduces to a
direct substitution into $f$ with no auxiliary construction.

\paragraph{Easy~/~Vector~\&~Coordinate.}
\textit{Problem.} $A(0,0)$, $B(6,0)$, $C(2,4)$; $\vec{AD} = \vec{AC} -
\vec{AB}$; $M$ is the midpoint of $AC$. Find $|DM|$.\\
\textit{Why selected.} Coordinates of $D$ and $M$ follow from direct
vector arithmetic; $|DM|$ is one distance-formula evaluation.

\paragraph{Medium~/~Euclidean.}
\textit{Problem.} $AB = AC = 10$, $BC = 12$. $O$ is the circumcenter;
$P$ on the bisector of $\angle BAC$ has perpendicular distance $2$
to $AB$. Find the perpendicular distance from $P$ to $BC$.\\
\textit{Why selected.} Three composed ideas---isosceles symmetry, the
half-angle sine relation, and a coordinate-distance check---yield
the answer, exemplifying a three-theorem \textsc{Medium} chain.

\paragraph{Medium~/~Function~\&~Calculus.}
\textit{Problem.} $f(x) = ax^2$ ($a > 0$). $A, B$ on graph at $x=0, 2$,
with $|AB| = 2\sqrt{10}$. Compute $\int_0^2 (f'(x))^2\, dx$.\\
\textit{Why selected.} The solver first recovers $a$ from the chord
length, then evaluates $\int_0^2(2ax)^2\,dx$; two named tools chained
through one unknown.

\paragraph{Medium~/~Vector~\&~Coordinate.}
\textit{Problem.} $AB=5$, $BC=6$, $AC=7$; $M$ mid of $AC$;
$\vec{AP} = \vec{AB}+\vec{AC}$, $\vec{AQ} = \vec{AB}-\vec{AC}$.
Compute $|MP|^2 + |MQ|^2$.\\
\textit{Why selected.} Dot-product expansion plus substitution of
$\vec{AB}\cdot\vec{AC} = (AB^2 + AC^2 - BC^2)/2$ from the law of
cosines---two named theorems with two derived points.

\paragraph{Hard~/~Euclidean.}
\textit{Problem.} $AB=AC$. $M$ mid of $BC$. The bisector of $\angle ABC$
meets $AC$ at $D$ and $AM$ at $P$. Given $AP = 3PM$ and $AM=12$,
find $|BP|$.\\
\textit{Why selected.} Four interleaved insights are required:
isosceles symmetry, the angle-bisector theorem, a similarity argument
for the $3{:}1$ ratio, and Pythagoras in $\triangle BMP$.

\paragraph{Hard~/~Function~\&~Calculus.}
\textit{Problem.} $f$ on $[0,6]$ consists of segment $OP$ from the
origin and a circular arc from $P$ to $Q$ on the $x$-axis, with
$x_P = 3$, $x_Q = 6$. The arc lies on a circle tangent to the
$x$-axis at $Q$, and $OP$ is tangent to the arc at $P$. Compute the
average rate of change of $f$ on $[0,3]$.\\
\textit{Why selected.} Two simultaneous tangency conditions yield a
non-linear system whose solution gives the radius and the height of
$P$; the implicit tangent--radius perpendicular is a
\textsc{Hard}-level structural marker.

\paragraph{Hard~/~Vector~\&~Coordinate.}
\textit{Problem.} Parallelogram $ABCD$, $M$ mid of $BC$,
$AM \perp BC$, $|BD| = 2|AB|$, $|AB|=4$. Compute the area.\\
\textit{Why selected.} Two independent structural constraints are
combined with the parallelogram identity $|BD|^2 + |AC|^2 = 2(|AB|^2 +
|BC|^2)$ and a vector projection---three derived points and a
four-equation system.

\subsection{Set 2}
\label{app:shots:set2}

\paragraph{Easy~/~Euclidean.}
\textit{Problem.} Triangle $ABC$ with $AB=6$, $BC=8$, $AC=10$.
$D, E, F$ are midpoints of $BC, AC, AB$. $K$ on $EF$ with $EK=1$.
Find $|DK|$.\\
\textit{Why selected.} The midsegment theorem reduces the
configuration to a coordinate computation; one named theorem to reach
the answer.

\paragraph{Easy~/~Function~\&~Calculus.}
\textit{Problem.} $f(x) = x^2$ and $g(x) = 2x + 3$. Let $N$ be the
number of intersection points on $[-2, 4]$; let $T$ be the
$y$-intercept of the tangent to $f$ at $x=1$. Compute $N + T$.\\
\textit{Why selected.} Two lightweight sub-tasks (intersection count
and tangent $y$-intercept) summed; no theorem chaining, just
composition of two elementary checks.

\paragraph{Easy~/~Vector~\&~Coordinate.}
\textit{Problem.} Triangle $ABC$, $AB = 5$, $AC = 7$. $D$ is
constructed so that $ABDC$ is a parallelogram. $M$ mid of $BC$,
$|AM|=4$. Find $|BC|$.\\
\textit{Why selected.} The parallelogram diagonal relation
$|AD|^2 + |BC|^2 = 2(|AB|^2 + |AC|^2)$ with $|AD| = 2|AM|$ gives
$|BC|$ in one identity.

\paragraph{Medium~/~Euclidean.}
\textit{Problem.} Trapezoid $ABCD$ with $AB \parallel CD$, $AB = 8$,
$CD = 12$. The diagonals meet at $P$; a line through $P$ parallel to
$AB$ meets $AD$ at $M$ and $BC$ at $N$. Find $|MN|$.\\
\textit{Why selected.} Solver must recognise that $|MN|$ is the
harmonic mean of the parallel sides---two similarity chains plus an
algebraic reduction.

\paragraph{Medium~/~Function~\&~Calculus.}
\textit{Problem.} $f(x) = x^3$ on $[0, 3]$. Let $c$ satisfy
$\int_0^c f(t)\,dt = 4$ and $d$ satisfy $f''(d) = 6$. Find $c + d$.\\
\textit{Why selected.} Two independent calculus skills---definite
integration and second-derivative root---feed a single algebraic sum.

\paragraph{Medium~/~Vector~\&~Coordinate.}
\textit{Problem.} $ABEC$ is a parallelogram with $AE \perp BC$, $AB=5$,
$BC=6$. Find $|AE|$.\\
\textit{Why selected.} Perpendicular diagonals force a rhombus-like
constraint that combines with the parallelogram diagonal identity;
two-theorem chain with one derived point.

\paragraph{Hard~/~Euclidean.}
\textit{Problem.} Square $ABCD$ with side $20$. $E \in BC$, $F \in CD$
with $BE = DF$; $AE, AF$ meet diagonal $BD$ at $P, Q$. Given
$|PQ| = 5\sqrt{2}$, find $|EF|$.\\
\textit{Why selected.} A $45^\circ$ rotation pairs $E \leftrightarrow
F$; similarity relates $|PQ|$ to $BE$; a Pythagorean step then
delivers $|EF|$---four chained ideas including a hidden symmetry.

\paragraph{Hard~/~Function~\&~Calculus.}
\textit{Problem.} Parabola $y = x^2$. $P$ lies on it in the first
quadrant. The tangent at $P$ meets the axes at $A, B$; the midpoint
$M$ of $AB$ lies on $y = -x$. Find the area of $\triangle OAB$.\\
\textit{Why selected.} Parametrise $P = (t, t^2)$, derive the
tangent, impose the midpoint-on-line condition, solve for $t$, and
compute the area---a four-step symbolic pipeline.

\paragraph{Hard~/~Vector~\&~Coordinate.}
\textit{Problem.} Equilateral $\triangle ABC$ with side $6$. $B$ is
the midpoint of $CP$; $G$ is the centroid of $\triangle ABC$. Find
$|PG|$.\\
\textit{Why selected.} Express $P$ via $\vec{CP} = 2\vec{CB}$, place
$G = (A + B + C)/3$, and evaluate $|PG|^2$ through dot products---three
derived points with an equilateral-triangle identity.

\subsection{Set 3}
\label{app:shots:set3}

\paragraph{Easy~/~Euclidean.}
\textit{Problem.} Right triangle with $\angle ACB = 90^\circ$,
$AC = 6$, $BC = 8$. $AD$ bisects $\angle CAB$, $D \in BC$;
$DE \perp AB$, $E \in AB$. Find $|DE|$.\\
\textit{Why selected.} The angle-bisector theorem on the right
triangle combined with the equal-tangent identity---one named theorem
with two derived points.

\paragraph{Easy~/~Function~\&~Calculus.}
\textit{Problem.} $f(x) = x^2 - 6x + 8$. $P, Q$ on graph at $x = 1, 5$;
$R$ is the vertex of the parabola. Find the area of $\triangle PQR$.\\
\textit{Why selected.} Three substitutions plus the triangle area
formula---no theorem chaining, but three named points force explicit
bookkeeping.

\paragraph{Easy~/~Vector~\&~Coordinate.}
\textit{Problem.} Parallelogram $ABCD$; $E \in AB$ with $AE = 4$,
$F \in CD$ with $CF = 4$; $AB = 12$, $AD = 6$. Find $|EF|$.\\
\textit{Why selected.} Vector decomposition $\vec{EF} = \vec{EB} +
\vec{BC} + \vec{CF}$ collapses to one magnitude evaluation.

\paragraph{Medium~/~Euclidean.}
\textit{Problem.} Triangle $ABC$ has area $120$; $M$ mid of $BC$; $D$
constructed so that $ABMD$ is a parallelogram. $P$ is the intersection
of $AC$ and $DM$. Find the area of $\triangle ADP$.\\
\textit{Why selected.} Parallelogram construction, similarity between
$\triangle ADP$ and $\triangle ACM$, and an area-ratio reduction---a
three-theorem \textsc{Medium}.

\paragraph{Medium~/~Function~\&~Calculus.}
\textit{Problem.} $f(x) = x^2 - 2x - 3$. Find the area of the region
bounded by the curve and the $x$-axis.\\
\textit{Why selected.} Standard definite-integral area problem after
solving a quadratic for the roots---two tools chained through the
roots.

\paragraph{Medium~/~Vector~\&~Coordinate.}
\textit{Problem.} Parallelogram $ABCD$ with $\vec{AB} \cdot \vec{AD} =
0$ ($AB = 6$, $AD = 8$). $P \ne C$ such that $PBCD$ is a kite with
$PB = CB$, $PD = CD$. Find $|AP|$.\\
\textit{Why selected.} Orthogonality turns the parallelogram into a
rectangle; the kite constraint reflects $P$ over a diagonal; $|AP|$
follows by coordinate computation---three steps composed.

\paragraph{Hard~/~Euclidean.}
\textit{Problem.} Isosceles trapezoid $ABCD$ with $AB \parallel CD$,
$AB = 10$, $CD = 4$. The Euler line of $\triangle ABC$ is parallel to
$AB$. Find $|AD|$.\\
\textit{Why selected.} Euler-line parallelism produces a
centroid--circumcenter alignment condition that reduces to a
non-linear equation in $|AD|$; the Euler-line invocation itself is a
\textsc{Hard}-level structural marker.

\paragraph{Hard~/~Function~\&~Calculus.}
\textit{Problem.} $f(x) = |x^2 - 16|$. $A, B$ on graph at $x = -2, 5$;
$L$ is the tangent to $f$ at $B$; $C$ is the intersection of $L$ with
the $y$-axis; $M$ on segment $AC$ with $AM : MC = 2 : 3$. Find $y_M$.\\
\textit{Why selected.} Absolute-value branch selection, tangent
construction on the branch containing $B$, and a weighted
midpoint---four chained sub-problems.

\paragraph{Hard~/~Vector~\&~Coordinate.}
\textit{Problem.} Triangle with $AB = 6$, $AC = 8$, $\angle BAC =
60^\circ$; $AD$ bisects $\angle BAC$ with $|AD| = 10$; $P$ is the
midpoint of $BD$. Find $|CP|$.\\
\textit{Why selected.} The angle-bisector length formula combines
with a Stewart-style ratio on $BD$ and a dot-product projection for
$|CP|$---three named theorems interleaved.

\subsection{Discussion}
\label{app:shots:discussion}

Across the three sets, the target-tier matching rates reported in
Table~\ref{tab:qwen35plus_difficulty_match} remain consistently high within each
tier, indicating that the judge's calibration does not hinge on any
single choice of exemplars: replacing all nine items with a
pairwise-disjoint substitute yields matching rates that vary by only
a few percentage points. When adapting VeriGeo to a new curriculum,
practitioners may therefore freely substitute the above exemplars
with curriculum-specific ones, provided each tier's items satisfy
the same reasoning-depth convention described in the selection
criteria above.

\section{Details of the Seed-Conditioned Difficulty Modulation Evaluation}
\label{app:seed_difficulty_modulation}

We provide additional details for the seed-conditioned difficulty modulation experiment reported in
the main text. This experiment evaluates whether \ourM{} can modulate the difficulty of a generated
variant relative to a given MathVista seed while preserving the seed's underlying geometric theme.

\paragraph{Seed-conditioned generation setting.}
We use the same 100 MathVista source problems as in the seed-conditioned generation experiment.
For each seed problem, we construct two generation settings. In the \textsc{Harder} setting, the
generator is instructed to produce a variant that preserves the seed's geometric theme but requires
a higher level of reasoning. In the \textsc{Equivalent} setting, the generator is instructed to produce
a variant with comparable difficulty while still preserving the original theme. In both settings, the
generator is required to maintain the core geometric context of the seed rather than replacing it with
an unrelated problem.

\paragraph{Knowledge Similarity Evaluation}
We evaluate the knowledge similarity in Figure~\ref{fig:knowledge_distributions}(b) using an LLM-as-a-judge prompt. The prompt is shown below:
\begin{verbatim}
"""
You are a geometry concept similarity evaluator.
Given two lists of geometry concepts from paired problems, score how similar 
their concept distributions are. Consider semantic equivalence and overlap.
Return JSON only with:
  - similarity: float between 0 and 1 (1 = identical, 0 = no overlap).
"""
\end{verbatim}

\paragraph{Pairwise difficulty-relation judge.}
We provide the prompt used for pairwise difficulty comparison, which asks the judge to determine whether one question is harder than another (\texttt{PROMPT\_HARDER}) or whether their difficulties are equivalent (\texttt{PROMPT\_EQUIV}).

\begin{verbatim}
PROMPT_HARDER = """
You are an LLM-based pairwise difficulty-relation judge for geometry problems.

You are given a seed problem and a generated variant authored from that seed. 
The requested generation condition is NOT provided to you. Your task is to 
compare the generated variant against the seed problem only after generation.

[SEED] (image attached)
SEED question: {seed_question}
SEED choices: {seed_choices}

[GENERATED]
GENERATED problem statement:
{gen_problem}

GENERATED proof outline (numbered steps):
{gen_proof}

Decide whether the GENERATED problem is strictly HARDER than the SEED problem.

Base your judgment on:
- changes in the number of reasoning steps;
- the degree of concept composition;
- the need for auxiliary constructions;
- whether the solution introduces additional algebraic or geometric 
  dependencies beyond those in the seed.

Answer "harder" only if the generated problem is meaningfully harder than the seed.
Otherwise, answer "not_harder".

Return STRICT JSON only, with no markdown:
{{"answer": "harder" | "not_harder", "reasoning": "<=3 sentences
explaining the judgment"}}
"""
\end{verbatim}

\begin{verbatim}
PROMPT_EQUIV = """
You are an LLM-based pairwise difficulty-relation judge for geometry problems.

You are given a seed problem and a generated variant authored from that seed. The requested 
generation condition is NOT provided to you. Your task is to compare the generated 
variant against the seed problem only after generation.

[SEED] (image attached)
SEED question: {seed_question}
SEED choices: {seed_choices}

[GENERATED]
GENERATED problem statement:
{gen_problem}

GENERATED proof outline (numbered steps):
{gen_proof}

Decide whether the GENERATED problem has approximately the same difficulty as the SEED problem.

Base your judgment on:
- changes in the number of reasoning steps;
- the degree of concept composition;
- the need for auxiliary constructions;
- whether the solution introduces additional algebraic or geometric 
  dependencies beyond those in the seed.

Answer "equivalent" if the generated problem is approximately the same difficulty as the 
seed. Answer "not_equivalent" if it is meaningfully harder or meaningfully easier.

Return STRICT JSON only, with no markdown:
{{"answer": "equivalent" | "not_equivalent", "reasoning": "<=3 sentences 
explaining the judgment"}}
"""
\end{verbatim}

\paragraph{Metric.}
Let $r_i \in \{\textsc{Harder}, \textsc{Equivalent}\}$ denote the requested relation for the $i$-th
seed-conditioned variant, and let $\hat r_i$ denote the relation predicted by the pairwise judge. We
define the target-difficulty matching rate as $\frac{1}{N}
\sum_{i=1}^{N}
\mathbf{1}\!\left[\hat r_i = r_i\right],$ where $N$ is the number of generated variants in the corresponding setting. A \textsc{Harder}
variant is counted as a match only when the judge determines that it is harder than the seed. An
\textsc{Equivalent} variant is counted as a match only when the judge determines that it has
approximately the same difficulty as the seed; variants judged as either easier or harder are treated
as mismatches.

\paragraph{Discussion.}
As reported in the main text, \ourM{} achieves a 100.0\% target-difficulty matching rate in the
\textsc{Harder} setting and an 80.0\% matching rate in the \textsc{Equivalent} setting. This gap is
expected. Increasing difficulty provides a clearer control direction, since the generator can introduce
additional reasoning steps, auxiliary constructions, or concept composition. By contrast, preserving
an exactly equivalent difficulty level is intrinsically more ambiguous, especially for seed problems
near the boundary between two difficulty levels. These judge-based results complement the structural
analysis in Figure~\ref{fig:knowledge_distributions}(c), where the difficulty-increased variants shift
toward higher concept counts while the equivalent-difficulty variants remain closer to the original
seed distribution.

\end{document}